\documentclass[runningheads]{llncs}
\pdfoutput=1  % arXiv: force PDFLaTeX so .pdf figures are accepted

\usepackage{eccv}

\usepackage[accsupp]{axessibility}
\usepackage[T1]{fontenc}
\usepackage{multirow}
\usepackage{colortbl}
\usepackage{nicefrac}
\usepackage[acronym]{glossaries}
\usepackage{titlecaps}  %
\usepackage{algorithm}
\usepackage{algpseudocode}
\usepackage{layout} %
\usepackage{import} %
\usepackage{color} %
\usepackage[most]{tcolorbox} %

\newcounter{prompt}
\renewcommand{\theprompt}{A.\arabic{prompt}}
\newtcolorbox{promptbox}[1]{%
  breakable, enhanced, colback=gray!5, colframe=black!60,
  fonttitle=\bfseries\small, coltitle=white, colbacktitle=black!60,
  fontupper=\small,
  left=3mm, right=3mm, top=2mm, bottom=2mm,
  title={Prompt~\theprompt: #1},
  sharp corners=downhill, boxrule=0.4pt
}
\usepackage{tikz}
\usepackage{tabularray}
\UseTblrLibrary{booktabs}
\usepackage{anyfontsize}
\usepackage{enumitem}
\usepackage{cuted}
\usepackage{etoolbox} %
\usepackage{chngcntr} %

\definecolor{plot_green}{HTML}{78b152}
\definecolor{plot_red}{HTML}{cd5f86}
\definecolor{plot_blue}{HTML}{7098d0}

\newcommand{\bs}[1]{\boldsymbol{#1}}
\newcommand{\para}[1]{\noindent\textbf{#1}}
\newcommand{\subpara}[1]{\noindent\textit{#1}}
\makeatletter
\newcommand{\dcref}[2]{%
  \def\dc@missing{0}%
  \forcsvlist{\dc@check}{#1}%
  \ifnum\dc@missing=1
    #2%
  \else
    \cref{#1}%
  \fi
}

\newcommand{\dc@check}[1]{%
  \@ifundefined{r@#1}{%
    \def\dc@missing{1}%
  }{}%
}
\makeatother
\newsavebox{\autoresizebox}
\newcommand{\autoresizetable}[1]{%
  \sbox{\autoresizebox}{#1}%
  \ifdim\wd\autoresizebox>\linewidth
    \resizebox{\linewidth}{!}{\usebox{\autoresizebox}}%
  \else
    \usebox{\autoresizebox}%
  \fi
}

\makeatletter
\newcommand\thefontsize[1]{{#1 The current font size is: \f@size pt\par}}
\makeatother

\newcommand{\name}{TriFlow} %
\newcommand{\best}[1]{\textbf{#1}} %
\newcommand{\second}[1]{#1} %

\newcommand{\glsfull}[1]{\glsentrylong{#1} (\glsentryshort{#1})}

\newacronym{qem}{QEM}{quadric error metric}
\newacronym{nvf}{NVF}{nearest-vertex vector field}
\newacronym{sdf}{SDF}{signed distance field}
\newacronym{tsdf}{TSDF}{truncated signed distance field}
\newacronym{fid}{FID}{Fr\'{e}chet Inception Distance}
\newacronym[
    plural=VLMs,
    longplural={vision-language models}
]{vlm}{VLM}{vision-language model}
\newacronym{vae}{VAE}{variational autoencoder}
\newacronym[
    plural=LODs,
    longplural={levels-of-detail}
]{lod}{LOD}{level-of-detail}

\usepackage{eccvabbrv}

\usepackage{graphicx}
\usepackage{booktabs}

\usepackage[accsupp]{axessibility}  %

\usepackage{hyperref}

\usepackage{orcidlink}

\begin{document}

\crefname{algorithm}{Alg.}{Algs.}
\crefname{prompt}{Prompt}{Prompts}
\Crefname{prompt}{Prompt}{Prompts}

\title{\name{}: Generating Artist-Like 3D Mesh Topology via Nearest-Vertex Vector Fields} 

\titlerunning{\name{}}

\author{Haoxuan Li\inst{1} \and
Ziya Erko\c{c}\inst{1} \and
Daniele Sirigatti\inst{2} \and
Vladislav Rosov\inst{2} \and
Lei Li\inst{3} \and
Angela Dai\inst{1} \and
Matthias Nie\ss{}ner\inst{1}}

\authorrunning{H.~Li et al.}

\institute{$^{1}$Technical University of Munich, Germany \quad $^{2}$AUDI AG, Germany \\ $^{3}$University of Virginia, USA}

\maketitle

\begin{figure}[h]
  \centering
  \sffamily
  \def\svgwidth{\linewidth}
  %% Creator: Inkscape 1.4.2 (f4327f4, 2025-05-13), www.inkscape.org
%% Accompanies image file 'teaser.pdf' (pdf, eps, ps)
\begingroup%
  \makeatletter%
  \providecommand\color[2][]{%
    \errmessage{(Inkscape) Color is used for the text in Inkscape, but the package 'color.sty' is not loaded}%
    \renewcommand\color[2][]{}%
  }%
  \providecommand\transparent[1]{%
    \errmessage{(Inkscape) Transparency is used (non-zero) for the text in Inkscape, but the package 'transparent.sty' is not loaded}%
    \renewcommand\transparent[1]{}%
  }%
  \providecommand\rotatebox[2]{#2}%
  \newcommand*\fsize{\dimexpr\f@size pt\relax}%
  \newcommand*\lineheight[1]{\fontsize{\fsize}{#1\fsize}\selectfont}%
  \ifx\svgwidth\undefined%
    \setlength{\unitlength}{365.23190308bp}%
    \ifx\svgscale\undefined%
      \relax%
    \else%
      \setlength{\unitlength}{\unitlength * \real{\svgscale}}%
    \fi%
  \else%
    \setlength{\unitlength}{\svgwidth}%
  \fi%
  \global\let\svgwidth\undefined%
  \global\let\svgscale\undefined%
  \makeatother%
  \begin{picture}(1,0.42595112)%
    \lineheight{1}%
    \setlength\tabcolsep{0pt}%
    \put(0,0){\includegraphics[width=\unitlength,page=1]{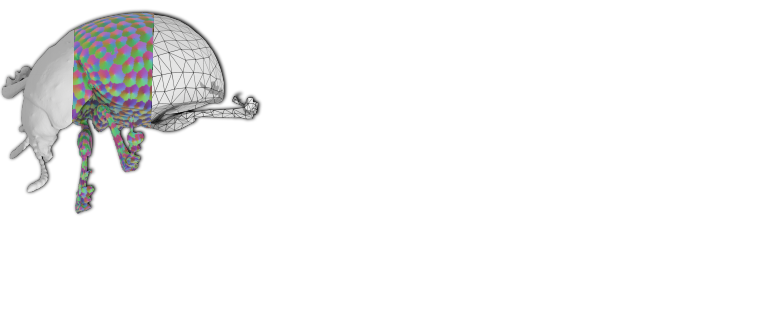}}%
    \put(0.05763473,0.36583543){\rotatebox{40.82717726}{\makebox(0,0)[t]{\lineheight{1.51285827}\smash{\begin{tabular}[t]{c}\fontsize{6pt}{4pt}\selectfont Geometry\end{tabular}}}}}%
    \put(0.14468004,0.41029172){\rotatebox{10.06343194}{\makebox(0,0)[t]{\lineheight{1.51285827}\smash{\begin{tabular}[t]{c}\fontsize{6pt}{4pt}\selectfont NVF\end{tabular}}}}}%
    \put(0.24712012,0.39815678){\rotatebox{-33.78509432}{\makebox(0,0)[t]{\lineheight{1.51285827}\smash{\begin{tabular}[t]{c}\fontsize{6pt}{4pt}\selectfont Mesh\end{tabular}}}}}%
    \put(0,0){\includegraphics[width=\unitlength,page=2]{teaser.pdf}}%
  \end{picture}%
\endgroup%

  \caption{\name{} generates high-quality, artist-like mesh topology from input SDFs. \textbf{Top~Left}: we introduce a \glsfull{nvf} to transform discrete topology prediction into efficient, piecewise-continuous field modeling. \textbf{Right}: \name{} robustly generalizes across diverse and complex geometries.}
  \label{fig:teaser}
\end{figure}

\begin{abstract}
We present \name{}, a new generative approach for producing compact 3D meshes with artist-like triangle topology directly from input geometry conditions such as signed distance fields.
Our key insight is to represent mesh topology as a \gls*{nvf} defined over the surface, where each point encodes its association to the nearest triangle vertex in the local barycentric frame. 
We train a latent flow-matching model to synthesize this field, enabling topology generation conditioned on the input geometry.
To extract a coherent mesh, we cluster surface regions using the generated \gls*{nvf} and guide a constrained \glsentrylong{qem} mesh simplification with topology-aware optimization.
This yields output meshes that closely match the input geometry while exhibiting structured, artist-like connectivity. 
Experiments demonstrate that \name{} achieves stronger generalization and significantly improved topology quality compared to state-of-the-art learning-based approaches, alongside 90\% lower Chamfer Distance and an 8$\times$ speedup.
\end{abstract}

\section{Introduction}
\label{sec:intro}

Triangle meshes serve as a fundamental representation across computer graphics and vision, broadly used in surface-based modeling, rendering, animation, physical simulation, 3D reconstruction, digital content creation, and beyond.
This prevalence stems from their computational efficiency, explicit surface structure, and direct compatibility with existing production tools. 
Beyond geometric accuracy, the \emph{topology} of a mesh plays a crucial role: the organization of vertices, edges, and faces strongly influences performance and applicability in downstream applications such as deformation, editing, and simulation;  poorly organized connectivity often leads to artifacts and costly manual cleanup. 
In practice, artists construct meshes with carefully designed topology, which we refer to as \emph{artist-like topology}, characterized by compact face count, smooth vertex distribution, and edge alignment with salient geometric features.

Recent works in 3D generative modeling~\cite{xiang2025trellis,xiang2025trellis2,chen2025dora,wu2025direct3ds2,he2025sparseflex,li2025triposg,ye2025hi3dgen,li2024craftsman3d} demonstrate impressive advances in geometry creation. 
However, the discrete nature of the optimized topologies remains difficult to synthesize directly.
These approaches produce high-fidelity geometry as implicit fields, requiring methods like Marching Cubes~\cite{lorensen1998marchingcubes} for mesh extraction, resulting in highly over-tessellated output meshes that are ill-suited for downstream workflows (e.g., real-time game rendering). To address this limitation, recent works have explored 
generating artist-like meshes directly as discrete triangle sequences in an autoregressive manner~\cite{siddiqui2024meshgpt,chen2025meshanythingv2,tang2024edgerunner,li2025meshpad,gao2025meshart,hao2024meshtron,lionar2025treemeshgpt,zhao2025deepmesh,xu2025meshmosaic,wang2025nautilus,zhang2025vertexregen}.
While such methods demonstrate promising results, they suffer from slow token-by-token inference and error accumulation during autoregressive prediction, which often leads to incomplete or low-fidelity generations -- limiting their scalability and generalization.
A significant gap still exists between high-fidelity 3D geometry modeling and robust, efficient, production-ready mesh topology generation.

To address this, we propose \name{}, a novel generative approach for creating meshes with artist-like topology.
Our key insight is that mesh topology can be represented as a piecewise-continuous
field defined over the surface, rather than as discrete connectivity. 
We thus introduce a \glsfull{nvf} that has a bijective mapping to the mesh topology.
This representation transforms the task of topology generation into vector-field modeling, thereby avoiding  expensive autoregressive sequence modeling. As illustrated in \cref{fig:teaser}, given input geometry encoded as a \gls*{sdf}, we predict the \gls*{nvf} and then extract a mesh with topology aligned to the field prediction.
Our approach is flexible and can handle any inputs that can be represented as \glspl*{sdf}, such as implicit fields from 3D generative methods~\cite{xiang2025trellis,park2019deepsdf}, and fused signed distance grids from sensor measurements~\cite{curless1996volumetric}. 

To recover meshes from the generated field, we develop a topology-aware extraction method that first clusters surface regions associated with respective target vertex positions using a watershed algorithm and then performs constrained \gls*{qem} simplification~\cite{garland1997qem}
guided by the predicted targets. This formulation not only preserves geometric fidelity but also effectively 
captures the topological characteristics
in our generated \gls*{nvf}. 
To enhance robustness against noisy or irregular input geometry commonly encountered in scanning and generative modeling, we introduce randomized surface distortions as data augmentation during training.

Our experiments demonstrate that \name{} significantly improves topology generation quality and generalization compared to the state of the art~\cite{xu2025meshmosaic,lionar2025treemeshgpt},
while achieving more than $8\times$ speedup.
By representing topology as a vector field, our method provides a new perspective on mesh generation that bridges geometry modeling and compact typology extraction.
To facilitate reproducibility, we release our complete codebase \footnote{Project page: \url{https://derkleineli.github.io/triflow/}}
for training and inference.

In summary, our contributions are:
\begin{itemize}[label=\textbullet, leftmargin=*, labelwidth=1.5em, labelsep=0.5em, topsep=0em, itemsep=0em, parsep=0em]
    \item We introduce a novel representation of mesh topology using a \glsfull{nvf} pointing towards the nearest vertex in the local triangle's barycentric frame.
    \item We propose a latent flow-matching approach for topology synthesis conditioned on input \gls*{sdf} geometry. We apply random surface distortion during training to enable the model to generalize to noisy or irregular geometry.
    \item We develop a robust topology-aware mesh extraction approach that combines a watershed algorithm with constrained QEM optimization to produce artist-like mesh outputs from our \gls*{nvf} prediction.
\end{itemize}

\section{Related Work}
\label{sec:relatedwork}

\para{3D Generation and Reconstruction.}
3D generation and reconstruction are highly demanded in practice.
Recent generative methods often represent geometry in implicit fields or structured 3D latents, including TRELLIS~\cite{xiang2025trellis,xiang2025trellis2}, Direct3D-S2~\cite{wu2025direct3ds2}, DORA~\cite{chen2025dora}, and recent high-resolution field models~\cite{he2025sparseflex,li2025triposg,ye2025hi3dgen,li2024craftsman3d}.
Similarly, reconstruction spans classical volumetric fusion into TSDF/SDF volumes~\cite{curless1996volumetric,newcombe2011kinectfusion,niessner2013voxelhashing} and neural field modeling such as NeRF~\cite{mildenhall2020nerf}, NeuS~\cite{wang2021neus}, and 3D Gaussian Splatting~\cite{kerbl2023gaussians}.
Many approaches therefore represent geometry as SDF/occupancy fields or related implicit formulations~\cite{park2019deepsdf,mescheder2019occupancy}.
Meshes are commonly recovered via iso-surfacing~\cite{lorensen1998marchingcubes,ju2002dualcontouring}
or via reconstruction methods~\cite{kazhdan2006poisson,kazhdan2013screenedpoisson}.
Differentiable and learned extraction variants further improve fidelity~\cite{shen2021dmtet,chen2021neuralmarchingcubes,shen2023flexicubes,chen2022neuraldualcontouring}, yet the resulting meshes are often dense and irregular and do not exhibit production-ready topology for editing and real-time rendering.

\para{Mesh Simplification and Retopology.}
Given dense proxy meshes, classical geometry processing reduces complexity via edge-collapse decimation, most notably QEM~\cite{garland1997qem} and Progressive Meshes~\cite{hoppe1996progressive}.
Several works adapt QEM to achieve the expected collapse behavior required by specific use cases~\cite{calderon2017boundingproxies,zhao2023variationalqem,salinas2015structureqem,trettner2020fastrobustqem}.
However, these approaches do not address the need for artist-like topology.
On the other hand, production retopology and quadrangulation use direction/cross fields to drive edge flow and patch layout, including N-RoSy design~\cite{palacios2007rosy}, globally optimal direction fields~\cite{knoppel2013directionfields}, MIQ~\cite{bommes2009miq}, Instant Field-Aligned Meshes~\cite{jakob2015instantmeshes}, QuadriFlow~\cite{huang2018quadriflow}, and spectral quadrangulation~\cite{dong2006spectralquadrangulation}.
More recently, learning-based approaches predict or generate such meshing fields, including learned direction fields~\cite{dielen2021learningdirectionfields}, CrossGen~\cite{dong2025crossgen}, and NeuFrameQ~\cite{liu2025neuframeq}.
These quad-meshing pipelines typically favor near-uniform, grid-like layouts, which diverge from artist-like topologies that allocate polygons adaptively to geometric details.

\para{Topology-Aware Mesh Generation.}
A strong recent trend is to model meshes as discrete sequences. 
These methods, including PolyGen~\cite{nash2020polygen}, MeshGPT~\cite{siddiqui2024meshgpt}, MeshAnything V2~\cite{chen2025meshanythingv2}, 
TreeMeshGPT~\cite{lionar2025treemeshgpt}, Nautilus~\cite{wang2025nautilus}, 
DeepMesh~\cite{zhao2025deepmesh}, and EdgeRunner~\cite{tang2024edgerunner}, Meshtron~\cite{hao2024meshtron},
generate meshes with autoregressive sequence modeling pipelines. 
Editing, scaling, and LOD-oriented systems include MeshPad~\cite{li2025meshpad}, 
MeshMosaic~\cite{xu2025meshmosaic}, 
and VertexRegen~\cite{zhang2025vertexregen}. 
While these methods are typically limited by the latency and error accumulation 
of autoregressive inference, our method bypasses this issue by modeling topology as the \gls*{nvf}.

\para{Flow Matching and Watershed for Surface Segmentation.}
Flow/rectified-flow objectives provide efficient, simulation-free training and fast sampling in continuous latent spaces~\cite{lipman2022flowmatching,liu2022rectifiedflow,tong2023conditionalflowmatching}, and have been adopted in large-scale 3D latent generators~\cite{xiang2025trellis,wu2025direct3ds2}.
For field-guided surface segmentation, watershed-style clustering is a classical segmentation tool for 2D images~\cite{vincent1991watersheds,cousty2009watershedcuts} and is extended to 3D mesh surfaces~\cite{mangan1999watershedmesh}.
Our method generates the \gls*{nvf} via flow matching and obtains surface clusters using watershed, which serve as topological constraints
in the mesh extraction.

\section{Method}
\begin{figure}[t]
  \centering
  \sffamily
  \def\svgwidth{\linewidth}
  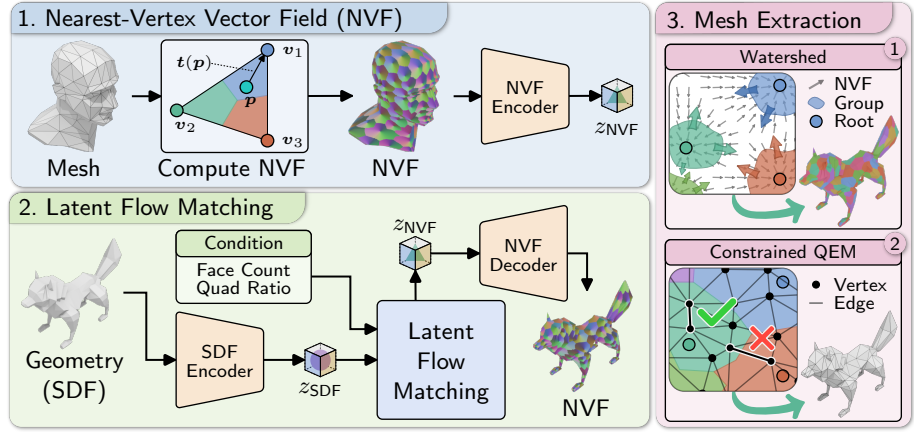
  \caption{Method Overview. Our method consists of three major components: \textbf{1}\onedot~We define the \glsentryshort{nvf} on the surface (local field directions color-coded) to represent the mesh topology;
  \textbf{2}\onedot~We train a latent flow-matching network to synthesize the \gls*{nvf} conditioned on the \glsentryshort{sdf} input; \textbf{3}\onedot~We use the watershed algorithm to cluster the surface and a constrained \glsentryshort{qem} that only merges vertices in the same group to extract the artist-like mesh as output. 
  }
  \label{fig:methodoverview}
\end{figure}

We introduce a novel generative approach to create compact, artist-like mesh topologies from \glsfull{sdf} inputs.
The \gls*{sdf}, denoted as $\mathcal{G}$, is used as the input geometry condition for its flexibility and ease of conversion from other representations. 
The surface, defined as the zero-level set of the SDF, is represented as $\mathcal{S} \subset \mathbb{R}^3$ containing all surface points.

Our goal is to learn a mapping from $\mathcal{G}$ to an artist-reminiscent mesh $\mathcal{M} = (\mathcal{V}, \mathcal{F})$, where $\mathcal{V} = \{\bs{v}_i \in \mathbb{R}^3 \mid i=1,\dots,N\}$ is the set of vertices, and $\mathcal{F} = \{f_j = (\bs{v}_{j_1}, \bs{v}_{j_2}, \bs{v}_{j_3}) \mid j=1,\dots,M\}$ is the set of triangle faces.
To efficiently predict the discrete mesh topology, we introduce a \glsfull{nvf} to represent mesh connectivity,
denoted as $\mathcal{T} = \{ \bs{t}(\bs{p}) \in \mathbb{R}^3 \mid  \forall \bs{p} \in \mathcal{S} \}$. 
Our mesh topology generation then follows a two-step pipeline: first generate $\mathcal{T}$ conditioned on $\mathcal{G}$; and then extract mesh $\mathcal{M}$ from the generated $\mathcal{T}$.

An overview of our method is illustrated in \cref{fig:methodoverview}. We train a latent flow-matching model to generate the \gls*{nvf} $\mathcal{T}$, conditioned on the \gls*{sdf} latent of the input geometry $\mathcal{G}$. 
To extract mesh topology, the generated field $\mathcal{T}$ is first clustered via a watershed algorithm to identify regions associated with target mesh vertices.
These regions are then used as constraints in the \gls*{qem} simplification method~\cite{garland1997qem} applied to a proxy mesh representing $\mathcal{G}$ to produce the final output mesh $\mathcal{M}$. These stages of the pipeline serve distinct, complementary roles. The latent flow-matching is the generative process. The generated \gls*{nvf} $\mathcal{T}$ encodes the complete target topology, including vertex positions and triangle formations. The watershed algorithm and the constrained QEM serve as a topology-extraction process, without introducing new topology information, providing robustness that absorbs prediction noise and voxelization aliasing with minimal hyperparameter tuning.

\subsection{\titlecap{\glsfull{nvf}}}
\label{sec:nvf}

\begin{figure}[tb]
  \centering
  \sffamily
  \begin{subfigure}[t]{0.3\linewidth}
    \def\svgwidth{\linewidth}
    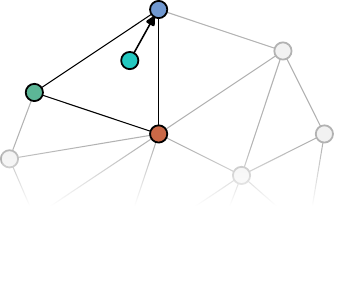
    \caption{Definition}
    \label{fig:nvf_def}
  \end{subfigure}
  \quad
  \begin{subfigure}[t]{0.3\linewidth}
    \def\svgwidth{\linewidth}
    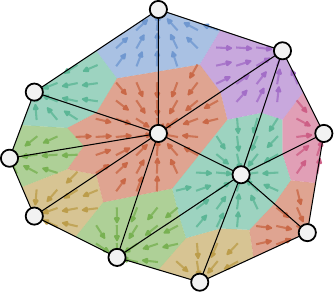
    \caption{Surface \glsentryshort{nvf}}
    \label{fig:nvf_surface}
  \end{subfigure}
  \quad
  \begin{subfigure}[t]{0.3\linewidth}
    \def\svgwidth{\linewidth}
    %% Creator: Inkscape 1.4.2 (f4327f4, 2025-05-13), www.inkscape.org
%% Accompanies image file 'sub3.pdf' (pdf, eps, ps)
\begingroup%
  \makeatletter%
  \providecommand\color[2][]{%
    \errmessage{(Inkscape) Color is used for the text in Inkscape, but the package 'color.sty' is not loaded}%
    \renewcommand\color[2][]{}%
  }%
  \providecommand\transparent[1]{%
    \errmessage{(Inkscape) Transparency is used (non-zero) for the text in Inkscape, but the package 'transparent.sty' is not loaded}%
    \renewcommand\transparent[1]{}%
  }%
  \providecommand\rotatebox[2]{#2}%
  \newcommand*\fsize{\dimexpr\f@size pt\relax}%
  \newcommand*\lineheight[1]{\fontsize{\fsize}{#1\fsize}\selectfont}%
  \ifx\svgwidth\undefined%
    \setlength{\unitlength}{161.56430054bp}%
    \ifx\svgscale\undefined%
      \relax%
    \else%
      \setlength{\unitlength}{\unitlength * \real{\svgscale}}%
    \fi%
  \else%
    \setlength{\unitlength}{\svgwidth}%
  \fi%
  \global\let\svgwidth\undefined%
  \global\let\svgscale\undefined%
  \makeatother%
  \begin{picture}(1,0.87010352)%
    \lineheight{1}%
    \setlength\tabcolsep{0pt}%
    \put(0,0){\includegraphics[width=\unitlength,page=1]{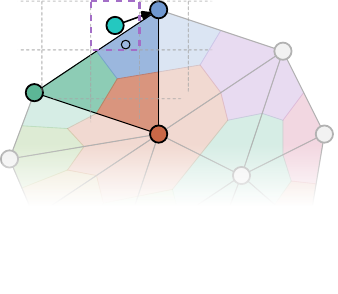}}%
    \put(0.55918486,0.82491284){\makebox(0,0)[t]{\lineheight{0.77810687}\smash{\begin{tabular}[t]{c}\fontsize{6pt}{0pt} $\bs{v}_1$\end{tabular}}}}%
    \put(0.61075001,0.67930325){\makebox(0,0)[t]{\lineheight{0.77810866}\smash{\begin{tabular}[t]{c}\fontsize{6pt}{0pt} $\bs{t}_c(\bs{p}_c)$\end{tabular}}}}%
    \put(0.37161842,0.67020681){\makebox(0,0)[t]{\lineheight{0.77810818}\smash{\begin{tabular}[t]{c}\fontsize{6pt}{0pt} $\bs{p}$\end{tabular}}}}%
    \put(0.18684286,0.78694468){\makebox(0,0)[t]{\lineheight{0.77810818}\smash{\begin{tabular}[t]{c}\fontsize{6pt}{0pt} $\bs{p}_c$\end{tabular}}}}%
    \put(0.49485866,0.07761721){\makebox(0,0)[t]{\lineheight{0.77810484}\smash{\begin{tabular}[t]{c}\scriptsize $\bs{t}_c(\bs{p}_c)=\bs{v}_1-\bs{p}_c$\end{tabular}}}}%
    \put(0,0){\includegraphics[width=\unitlength,page=2]{sub3.pdf}}%
  \end{picture}%
\endgroup%

    \caption{Sparse Voxelization}
    \label{fig:nvf_voxelization}
  \end{subfigure}
  \caption{Definition of the \glsfull{nvf}. \textbf{(a)} The \glsentryshort{nvf} is defined as a vector pointing towards the vertex with the largest barycentric weight. \textbf{(b)} The \glsentryshort{nvf} defines surface regions whose connectivity aligns with the vertex connectivity. \textbf{(c)} We voxelize the \glsentryshort{nvf} for neural network training.
  }
  \label{fig:nvf}
\end{figure}

The \gls*{nvf} $\mathcal{T}$ is a vector field defined on the mesh $\mathcal{M}$, as illustrated in \cref{fig:nvf_def}. Given any surface point $\bs{p} \in \mathcal{S}$, the field value $\bs{t}(\bs{p})$ is the vector pointing towards the nearest mesh vertex of $\bs{p}$ in the local triangle's barycentric frame:
$\bs{t}(\bs{p}) = \bs{v}_n - \bs{p}$,
where $\bs{v}_n$ denotes the nearest vertex. Given the triangle face $f_n=(\bs{v}_{1}, \bs{v}_{2}, \bs{v}_{3})$  containing the point $\bs{p}$, we build a barycentric coordinate system and represent $\bs{p}$ as
$\bs{p} = \lambda_1 \bs{v}_{1} + \lambda_2 \bs{v}_{2} + \lambda_3 \bs{v}_{3}$,
where $(\lambda_1,\lambda_2,\lambda_3)$ are the barycentric coordinates satisfying $\lambda_i \ge 0$ and $\lambda_1+\lambda_2+\lambda_3=1$. The nearest vertex index is determined by
\begin{equation}
    n = \arg\max_{i \in \{1,2,3\}} \lambda_i,
    \label{eq:barynn}
\end{equation}
which assigns $\bs{p}$ to the vertex with the maximal barycentric weight in the local triangle.

To enable efficient generative modeling, we discretize the surface $\mathcal{S}$ and the \gls*{nvf} $\mathcal{T}$ into a sparse voxel grid.
As shown in \cref{fig:nvf_voxelization}, for each voxel intersecting with $\mathcal{S}$, we take the voxel center $\bs{p}_c$ and find its nearest point $\bs{p}$ on the surface and the triangle $f_n$ containing $\bs{p}$. Then \cref{eq:barynn} is used to determine the nearest vertex $\bs{v}_n$. The voxelized \gls*{nvf} at this voxel is defined as
$\bs{t}_c(\bs{p}_c) = \bs{v}_n - \bs{p}_c$.

This representation partitions each triangle into three regions corresponding to its incident vertices, implicitly encoding topology.
Points mapped to the same triangle vertex form a connected surface region, with region adjacency corresponding to the mesh vertex connectivity, as shown in \cref{fig:nvf_surface}.
Unlike the discrete vertex-face representation $(\mathcal{V}, \mathcal{F})$, the \gls*{nvf} is piecewise continuous and suitable for neural modeling.

\subsection{Latent Flow-Matching Model for NVF}
\label{sec:latentflow}

We learn to generate mesh topology by predicting the latent representation of the \gls*{nvf} $\mathcal{T}$, conditioned on the input \gls*{sdf} $\mathcal{G}$. As shown in \cref{fig:methodoverview}, the model takes as input an \gls*{sdf} latent $z_{\text{SDF}}$ representing $\mathcal{G}$,
together with user-controlled topology parameters, and generates a latent $z_{\text{NVF}}$ to be decoded into the \gls*{nvf}.

\para{\glsentryshort{sdf} Condition Encoding.}
We voxelize the input \gls*{sdf} $\mathcal{G}$ on a $512^3$ grid and discard voxels with unsigned distances greater than $\nicefrac{1}{128}$ of the maximum extent of its bounding box, resulting in a sparse voxel representation. 
We train an autoencoder that encodes the \gls*{sdf} grid into a sparse latent grid $z_{\text{SDF}}$ of resolution $64^3$.
The model is trained with an $\ell_1$ reconstruction loss:
$\mathcal{L}_{\text{SDF}} = \| \hat{\text{SDF}} - \text{SDF} \|_1$.
The resulting latent serves as a compact geometric condition for generating mesh topology.

\para{\gls*{nvf} Encoding and Decoding.}
The \gls*{nvf} $\mathcal{T}$ is voxelized at resolution $512^3$ and encoded using a 
\gls*{vae} into a sparse latent grid $z_{\text{NVF}}$ of size $64^3$. 
We parameterize the field $\mathcal{T}$ using its unit direction $\bs{d} = \nicefrac{\mathcal{T}}{\|\mathcal{T}\|_2}$ and the square root of its magnitude $s = \sqrt{\|\mathcal{T}\|_2}$, which are the direct output of the decoder. This decomposition prioritizes directional consistency in low-magnitude regions near mesh vertices, which is critical for accurate mesh extraction.
We train the model using an $\ell_1$ loss on the reconstructed direction $\hat{\bs{d}}$ and scaled magnitude $\hat{s}$, alongside a KL-divergence loss weighted by $\lambda_{KL}$:
\begin{equation}
\mathcal{L}_{\text{NVF}} = \| \hat{\bs{d}} - \bs{d} \|_1 + \| \hat{s} - s \|_1 + \lambda_{KL}\mathcal{L}_{KL}.
\end{equation}

\para{Latent Flow Matching.}
Given the SDF latent $z_{\text{SDF}}$ and topology control parameters consisting of the target face count and quad-face ratio, which control the output mesh density and topological regularity,
we train a latent flow-matching network 
\cite{xiang2025trellis} to generate the \gls*{nvf} latent $z_{\text{NVF}}$. 
Training is performed along a linear interpolation path between a ground-truth latent $z_0$ and Gaussian noise $\epsilon$,
\begin{equation}
z(i) = (1 - i) z_0 + i\epsilon, \quad i \in [0,1],
\end{equation}
which progressively transforms data samples into noise. The network learns a time-dependent velocity field $u$ that moves samples along this path toward the data distribution. The training follows the conditional flow-matching objective:
\begin{equation}
\mathcal{L}_{\text{flow}} =
\mathbb{E}_{i,z_0,\epsilon}\left[
\left\| u_\theta(z(i),i,z_{\text{SDF}},c) - (\epsilon - z_0) \right\|_1
\right],
\end{equation}
where $c$ denotes the topology control parameters. The generated latent is finally decoded to obtain the voxelized \gls*{nvf} $\mathcal{T}$.

\subsection{Mesh Extraction from NVF}
\label{sec:extractmesh}

To mitigate clustering ambiguity caused by prediction uncertainty, we first process the predicted \gls*{nvf} to obtain a spatially smooth field, ensuring that surface points are robustly grouped into consistent components.
The topology intrinsic to the surface grouping is then transferred to a mesh as output. This process is also constrained by optimization objectives in accordance with \gls*{qem} for geometry alignment.

\para{\gls*{nvf} Smoothing and Transfer.}
Given the voxelized \gls*{nvf} prediction from the flow-matching model, we first apply a bilateral filter to suppress noise in locally smooth regions inferred by the network.
We extract an over-tessellated mesh $\mathcal{M}_d=(\mathcal{V}_d,\mathcal{F}_d)$ from the \gls*{sdf} $\mathcal{G}$ using marching cubes~\cite{lorensen1998marchingcubes} as a proxy whose vertices densely sample the surface of $\mathcal{G}$.
We then transfer the voxelized \gls*{nvf} onto the vertices of $\mathcal{M}_d$ as follows.
For each vertex $\bs{v}_d$ in $\mathcal{M}_d$, we locate its nearest voxel center $\bs{p}_{c_n}$ and assign an \gls*{nvf} vector according to
\begin{equation}
    \bs{t}_d(\bs{v}_{d}) = \bs{v}_{n} - \bs{v}_{d}, 
    \qquad
    \bs{v}_{n} = \bs{p}_{c_n} + \bs{t}_c(\bs{p}_{c_n}),
\end{equation}
where $\bs{t}_c(\bs{p}_{c_n})$ denotes the predicted \gls*{nvf} at the voxel center. This assignment ensures that the vector associated with $\bs{v}_d$ points toward the same target position indicated by the corresponding voxel prediction.

\para{Watershed Algorithm.}
Ideally, the NVF naturally groups surface points $\mathcal{S}$ into distinct clusters, where points in the same cluster share a common target vertex (\cref{fig:nvf_surface}). However, in practice, prediction noise and voxelization artifacts can cause clustering ambiguity.
To robustly recover the connected components defined by the \gls*{nvf} prediction, we introduce a watershed algorithm that iteratively expands regions on the surface of $\mathcal{M}_d$.
The algorithm consists of the following three steps:

\subpara{1) Region Root Initialization.}
Region roots are defined as vertices in $\mathcal{V}_d$ with small predicted displacement magnitudes, indicating they are spatially close to a target mesh vertex:
\begin{equation}
\mathcal{R} = \{\bs{r}\in\mathcal{V}_d \mid \|\bs{t}_d(\bs{r})\|_\infty \le \tau\},
\label{eq:roots}
\end{equation}
where $\tau$ is a threshold. 
Each root seeds an initial cluster.

\subpara{2) Iterative Region Expansion.}
Let $\bs{x}(\bs{v}_d) = \bs{v}_d + \bs{t}_d(\bs{v}_d)$ denote the predicted target position for a vertex $\bs{v}_d \in \mathcal{V}_d$.
Starting from the region roots, cluster labels are propagated over the mesh adjacency graph of $\mathcal{M}_d$ using a priority queue, iteratively growing the seed regions. 
At each step, an unlabeled neighboring vertex joins the cluster of the root $\bs{r} \in \mathcal{R}$ that offers the closest predicted target position in Euclidean distance:
\begin{equation}
\mathrm{cost}(\bs{v}_d, \bs{r})
= \left\| \bs{x}(\bs{v}_d) - \bs{x}(\bs{r}) \right\|_2 .
\label{eq:cost}
\end{equation}
Vertices are processed in ascending order of this cost, producing a watershed-like expansion that favors spatially coherent regions.

\subpara{3) Updating \gls*{nvf}.}
Once all vertices are assigned to clusters, we update the \gls*{nvf} on $\mathcal{M}_d$ according to the predicted position of region roots:
\begin{equation}
\bs{t}_d(\bs{v}_d) \gets \bs{x}(\bs{r}) - \bs{v}_d.
\label{eq:updatenvf}
\end{equation}

\para{Constrained \gls*{qem}.}
While the \gls*{nvf} captures the desired artist-like connectivity, the predicted vertex positions often exhibit small misalignments with $\mathcal{G}$. 
To achieve high geometric fidelity, we ``snap'' the topology to $\mathcal{G}$ via a geometry-aware optimization.
Therefore, we utilize the proxy mesh $\mathcal{M}_d$ for the geometry $\mathcal{G}$. Optimizing $\mathcal{M}_d$ by combining topology constraints derived from the \gls*{nvf} and the vanilla \gls*{qem} quadrics results in a mesh aligned to both the intended topology and the original geometry $\mathcal{G}$.
The vanilla \gls*{qem} simplifies a mesh by iteratively collapsing vertex pairs while maintaining surface error approximations using quadric matrices.
Each vertex is associated with a symmetric $4 \times 4$ matrix $Q_{\text{geom}}$ that encodes the incident triangle set.
Each edge is assigned a cost computed from the vertex quadrics, measuring the geometric distortion induced by collapsing the edge.
Edges with costs below a certain threshold are collapsed to obtain a simplified mesh while preserving the overall geometry.
However, the vanilla \gls*{qem} does not faithfully respect artist-like mesh topology.

Our key idea is to restrict edge collapses according to the predicted topology. Specifically, an edge collapse is rejected if the two vertices have different region roots, as determined by the watershed algorithm. This prevents undesirable merging of distinct regions and aligns the resulting topology with the NVF prediction.
When a valid edge collapse involves root vertices, we bias the resulting contraction point toward the corresponding vertex position predicted by the generative model.
We achieve this by augmenting the geometric quadric with an additional positional constraint $Q_t$.
The resulting quadric is defined as
\begin{equation}
    Q = \mathrm{mean}(Q_{\text{geom}}) + \lambda_t Q_t,\quad Q_t =
\begin{bmatrix}
1 & 0 & 0 & -x_t \\
0 & 1 & 0 & -y_t \\
0 & 0 & 1 & -z_t \\
-x_t & -y_t & -z_t & \|\bs{x}(\bs{v}_d)\|_2^2
\end{bmatrix},
\end{equation}
where $\mathrm{mean}(Q_{\text{geom}})$ denotes the average geometric quadrics accumulated at the two vertices of the edge, $Q_t$ is a quadratic penalty encouraging the vertex position to remain close to the target $\bs{x}(\bs{v}_d) = (x_t, y_t, z_t)^\intercal$, and $\lambda_t$ controls the constraint strength. 

\subsection{Geometric Data Augmentation}
To improve robustness against noise and irregularity common in generated or scanned geometry, we apply a random 3D distortion field to the training data. By randomly perturbing the geometry and its corresponding \gls*{nvf} fields, we force the model to learn stable topological predictions even in the presence of significant local surface noise or geometric distortions. More details, including the definition of the distortion field, are provided in the supplementary material.

\section{Experiments}
\label{sec:experiments}

\subsection{Implementation Details}
\label{sec:implementation_details}

The training dataset contains 656k Objaverse~\cite{deitke2023objaverse} samples. 
The \gls*{sdf} autoencoder is built from fine-tuning the VAE of Direct3D-S2~\cite{wu2025direct3ds2} for 170k iterations with a batch size of 16 on 8$\times$A6000 GPUs for 8 days.
The \gls*{nvf} \gls*{vae} adopts the same architecture, and is trained with $\lambda_{\text{KL}}=0.001$ for 252k iterations with batch size 32 on 8$\times$H100 GPUs for 11 days. 
The latent dimension of both \gls*{sdf} and \gls*{nvf} is $16$.
We build our conditional latent flow matching model based on the TRELLIS~\cite{xiang2025trellis} architecture.
The model is trained for 179k iterations with a batch size 64 on 8$\times$H100 GPUs for 12 days. 
All evaluations and runtime measurements are conducted on a single A6000 GPU with 4 CPU cores.
Please refer to the supplementary material for more details.

\subsection{Metrics and Baselines}
\label{sec:metricsandbaselines}

\para{Evaluation Metrics.}
We evaluate both geometric fidelity and topological quality using complementary metrics. 
Geometric accuracy is evaluated using Chamfer Distance between the input and output surfaces after normalizing meshes to a unit cube. We uniformly sample 10k surface points from each mesh to compute the distance.
We report \gls*{fid} between shaded renderings of generated and ground-truth meshes to quantify visual geometric similarity.
Additionally, we introduce perceptual ratings from \glspl*{vlm}~\cite{qwen35blog,team2023gemini} on mesh geometry and topology to evaluate the visual coherency with artist-created meshes. We also let \glspl*{vlm}  compare wireframe renderings and report pairwise preferences of our method against baseline methods. We further conduct a human perceptual study where participants rate and compare outputs from different approaches.

\para{Baselines.}
We compare against both classical mesh simplification methods and recent learning-based approaches. 
Classical baselines include vanilla \gls*{qem}~\cite{garland1997qem,zhou2018open3d} and QuadriFlow~\cite{huang2018quadriflow,blender}, which represent widely used geometry-driven mesh simplification and remeshing techniques. 
We also include TreeMeshGPT~\cite{lionar2025treemeshgpt} and MeshMosaic~\cite{xu2025meshmosaic} as state-of-the-art learning-based methods that model priors of artist-created topology. 
Together, these baselines enable a comprehensive evaluation of topology quality, geometric preservation, and generalization.

\subsection{Experimental Setup}

\para{Evaluation Data.}
We conduct evaluation on two datasets. The first consists of~$\sim 1000$ test samples from Objaverse~\cite{deitke2023objaverse}, chosen to guarantee single-component meshes with clean, artist-like topology. The second dataset contains $65$ shapes generated by TRELLIS~\cite{xiang2025trellis}, which exhibit challenging geometries with artifacts commonly observed in generative approaches, such as bumpy surfaces. 
Evaluation on TRELLIS data demonstrates the practicality and generalization of our method in real-world 3D content creation pipelines.

\para{Target Face Counts.}
For Objaverse samples, we set the target number of faces for all capable methods --- including Ours, QEM, and QuadriFlow --- to match the ground-truth mesh. For TRELLIS samples, we set the target to $\sim 6000$ faces.

\para{Method-Specific Settings.}
For the part-based baseline MeshMosaic~\cite{xu2025meshmosaic}, Objaverse shapes are treated as single-part meshes. Following their recommended pipeline, TRELLIS shapes are decomposed into $20$ parts using PartField~\cite{liu2025partfield}. QuadriFlow requires manifold meshes as input, and thus non-manifold samples are excluded when computing evaluation metrics for QuadriFlow to ensure a fair comparison. We use the target quad-face ratio of $0.95$, the root threshold $\tau$ as \nicefrac{1}{2} voxel size, and the topology weight $\lambda_t = 0.1$ in our method. The impact of these parameters is discussed in the supplementary material.

\subsection{Comparison to State of the Art}
\label{sec:experiment_comparison}

\begin{figure}[h!]
\centering
\sffamily
\def\svgwidth{\linewidth}
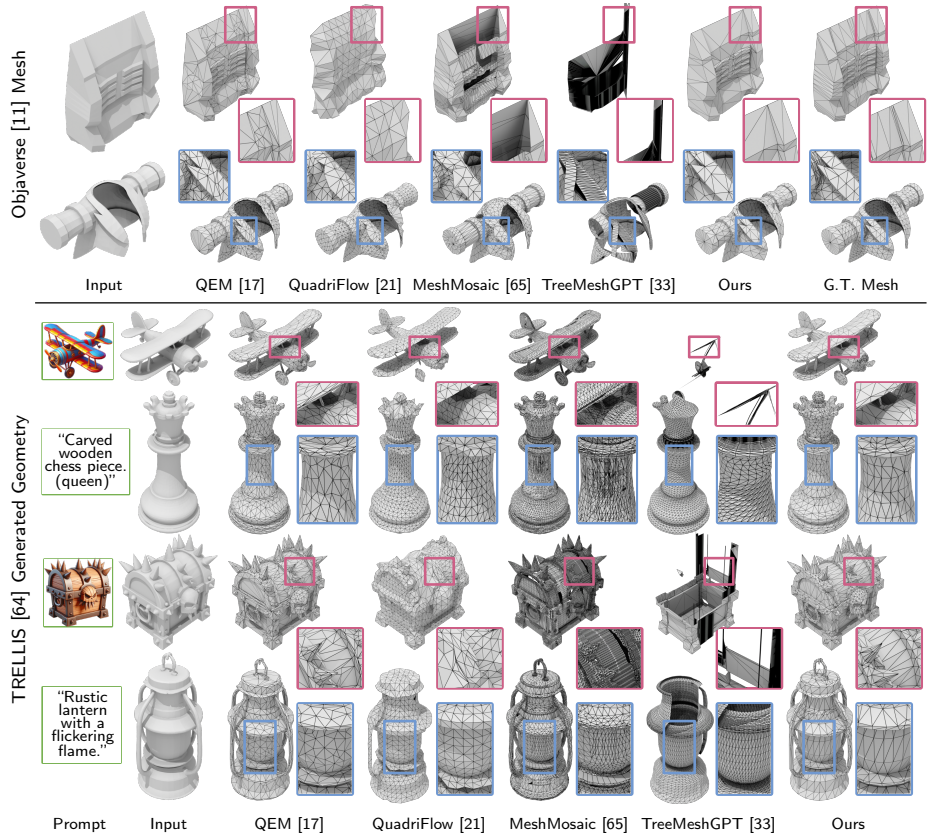
\caption{Visual Comparison. \name{} demonstrates superior robustness across diverse input geometries. We specifically evaluate on outputs from TRELLIS~\cite{xiang2025trellis} to simulate a practical production pipeline, where raw generative outputs with over-tessellated meshes require conversion into clean topology.
Our approach consistently produces high-quality, artist-like topologies characterized by compact triangles and regular edge flows, while maintaining high fidelity to the original geometry.}
\label{fig:visualcompare}
\end{figure}

\begin{table}[t]
\centering
\caption{Quantitative comparison with state of the art on Objaverse~\cite{deitke2023objaverse} and TRELLIS~\cite{xiang2025trellis}-generated shapes. Chamfer Distance is scaled by a factor of $1000$. $\{\text{M},\text{U}\}_{\{\mathcal{G},\mathcal{T}\}}$ represents perceptual scores of geometrical similarity and topological likeness to artist meshes provided by \glspl*{vlm}~(M) and human volunteers~(U), respectively. Our method consistently outperforms baselines in both geometric similarity and topological quality.
}
\label{tab:comparison}
\autoresizetable{ 
\SetTblrInner{rowsep=1pt,colsep=2pt}
\begin{tblr}{
  colspec = {l | *{4}{c} | *{6}{c}},
  row{1-2} = {font=\bfseries}, 
}
\toprule
 & \SetCell[c=4]{c} Objaverse~\cite{deitke2023objaverse} & & & & \SetCell[c=6]{c} TRELLIS~\cite{xiang2025trellis} Geometry & & & & & \\
Method & CD$\downarrow$ & FID$\downarrow$ & $\textbf{M}_\mathcal{G}$$\uparrow$ & $\textbf{M}_\mathcal{T}$$\uparrow$ & CD$\downarrow$ & FID$\downarrow$ & $\textbf{M}_\mathcal{G}$$\uparrow$ & $\textbf{M}_\mathcal{T}$$\uparrow$ & $\textbf{U}_\mathcal{G}$$\uparrow$ & $\textbf{U}_\mathcal{T}$$\uparrow$ \\
\midrule
QEM~\cite{garland1997qem}         & 0.14          & \second{8.0}  & \second{4.3}  & 3.3         & \best{0.20} & \second{22.1} & \second{3.6} & \second{3.3} & \second{4.0} & \second{3.6}  \\
QuadriFlow~\cite{huang2018quadriflow}  & 4.81          & 46.9          & 3.0           & 2.7         & 0.52        & 95.6          & 3.4          & 3.1          & 2.5 & 3.2  \\
MeshMosaic~\cite{xu2025meshmosaic}  & 1.62          & 27.0          & 3.9           & \second{3.6}& 9.02        & 45.3          & 2.8          & 1.7          & 3.3 & 2.1  \\
TreeMeshGPT~\cite{lionar2025treemeshgpt} & \second{0.98} & 8.1           & 3.8           & 3.5         & 30.00       & 74.9          & 1.9          & 1.7          & 1.5 & 1.7  \\
Ours        & \best{0.12}   & \best{5.0}    & \best{4.8}    & \best{4.7}  & \best{0.20} & \best{16.2}   & \best{4.4}   & \best{4.0}   & \best{4.5} & \best{4.6}  \\
\bottomrule
\end{tblr}
}
\end{table}

\begin{table}[t]
\centering
\caption{Perceptual preferences of our method compared to baselines, based on \glspl*{vlm} and human participants. Our method is consistently preferred in both geometry and topology quality.}
\label{tab:preference}
\autoresizetable{
\begin{tabular}{@{\extracolsep{5pt}}lccccc}
\toprule
\textbf{Ours vs.} & \textbf{QEM}~\cite{garland1997qem} & \textbf{QuadriFlow}~\cite{huang2018quadriflow} & \textbf{MeshMosaic}~\cite{xu2025meshmosaic} & \textbf{TreeMeshGPT}~\cite{lionar2025treemeshgpt} \\
\midrule
VLM-Geometry     & 70.7\% & 81.8\% & 93.0\% & 82.9\%      \\
VLM-Topology     & 65.9\% & 84.8\% & 74.4\% & 92.7\%      \\
VLM-Preference     & 68.3\% & 84.8\% & 76.7\% & 92.7\%      \\
Human-Geometry    & 76.5\% & 95.7\% & 84.4\% & 97.9\%       \\
Human-Topology    & 80.2\% & 94.9\% & 96.1\% & 95.0\%       \\
Human-Preference    & 83.3\% & 97.1\% & 90.8\% & 97.9\%       \\
\bottomrule
\end{tabular}
}
\end{table}

We compare \name{} against classical approaches and learning-based methods on Objaverse~\cite{deitke2023objaverse} and TRELLIS~\cite{xiang2025trellis}-generated shapes. 
Objaverse serves as an in-domain benchmark, as the learning-based methods are trained on its distribution. 
In contrast, TRELLIS-generated shapes provide out-of-distribution validation, serving as a practical test of converting over-tessellated 3D generations to compact, production-ready mesh structures.
Quantitative results are summarized in \cref{tab:comparison}, perceptual preference studies are reported in \cref{tab:preference}, and qualitative comparisons are shown in \cref{fig:visualcompare}.

Overall, \name{} consistently outperforms prior approaches: across TRELLIS shapes and Objaverse, our method obtains either the best or tied-best Chamfer Distance while simultaneously achieving the lowest \gls*{fid} and the highest perceptual geometry and topology scores. 
The preference study in \cref{tab:preference} further supports this trend, where both VLM-based evaluation and human judgments consistently favor meshes produced by \name{} across geometry, topology, and overall preference.

A notable observation is the limited generalizability of learning-based baselines when evaluated beyond their training distribution, denoted by their performance gap between Objaverse and TRELLIS shapes in \cref{tab:comparison}. We attribute this to two fundamental design limitations. First, these approaches represent meshes as ordered token sequences. As a result, a significant portion of model capacity is spent on modeling sequence ordering, reducing the effective capacity for modeling geometric and topological relations. Second, autoregressive generation introduces accumulated errors: under out-of-distribution inputs such as TRELLIS geometry, errors accumulate faster, leading to more failures, as reflected in \cref{fig:visualcompare}. In contrast, \name{} avoids sequential dependency during generation, enabling more stable predictions and consistent performance across both datasets.

\begin{figure}[t]
\centering
\sffamily

\def\svgwidth{\linewidth}
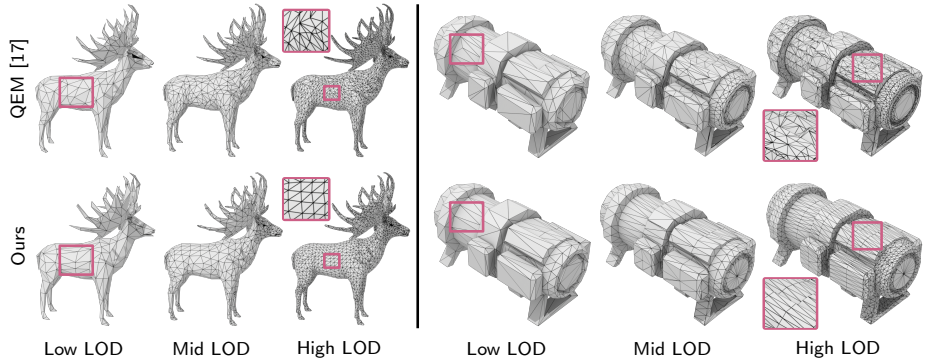
\caption{\glsentryshort{lod} comparison. Our method produces regular and efficient triangle layout across multiple \glsentryshortpl{lod}, while \glsentryshort{qem} results in irregular triangle soup with the same triangle count budget.
  }
\label{fig:lodresult}
\end{figure}

\para{Face Orientation.}
The dense proxy mesh is derived from the input \gls*{sdf} and is therefore normal-consistent; the constrained \gls*{qem} simplification preserves this consistency, so \name{} naturally produces meshes with consistent face orientation. On the other hand, autoregressive baselines~\cite{lionar2025treemeshgpt,xu2025meshmosaic} do not reliably generate triangles with consistent face normals.

\para{Level-of-Detail (\glsentryshort{lod}) Topology.}
\Gls*{lod} is required in modern graphics pipelines, where meshes must remain visually consistent under varying computational budgets and viewing distances. We therefore evaluate \name{}'s performance across multiple simplification levels and compare it with \gls*{qem}~\cite{garland1997qem}.
As shown in \cref{fig:lodresult}, \name{} excels in shape abstraction at low \glspl*{lod}, producing regular edge flows that mirror the desirable topology of artist-crafted low-poly models. At higher resolutions, \name{} yields a dense, structured tessellation characteristic of artist-like polygonal modeling. While \gls*{qem} offers competitive shape preservation, it produces highly irregular triangular meshes that may pose challenges to downstream editing and lead to artifacts in rendering pipelines.
In contrast, \name{} introduces a learned topology prior without compromising geometric fidelity, providing \gls*{lod} meshes that feature both well-preserved geometry and optimized topology.

\para{Runtime.} While autoregressive learnable baselines are capable of generating a large amount of triangles, they are fundamentally limited by the latency of token-by-token generation. Consequently, processing a single sample takes $2.2$ hours with MeshMosaic~\cite{xu2025meshmosaic} and $4.3$ minutes with TreeMeshGPT~\cite{lionar2025treemeshgpt}. In contrast, by representing topology as a vector field, \name{} achieves significantly accelerated generation: processing a sample takes only $31$ seconds.

\subsection{Ablations}

\begin{figure}[t]
  \centering
  \sffamily
  \begin{subfigure}[t]{0.4\linewidth}
    \centering
    \def\svgwidth{0.75\linewidth}
    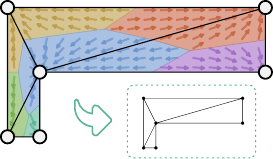
    \caption{Using Barycentric Coordinates}
  \end{subfigure}
  \quad
  \begin{subfigure}[t]{0.4\linewidth}
    \centering
    \def\svgwidth{0.75\linewidth}
    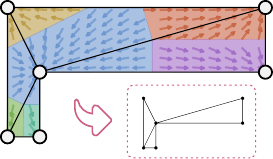
    \caption{Using Euclidean Distance}
  \end{subfigure}
  \caption{Analysis of using barycentric weights in \glsentryshort{nvf}. This simple example shows that defining a nearest-vertex vector field with  barycentric weights \textbf{(a)} faithfully represents the topology, compared to the naive  Euclidean distance \textbf{(b)}.
  }
  \label{fig:baryvsspacial}
\end{figure}

We ablate design choices and individual contribution components of our pipeline, covering \gls*{nvf} formulation, watershed grouping, \gls*{qem} optimization, and geometric data augmentation.

\para{Defining \gls*{nvf} with Barycentric Frames.}
We first analyze our \gls*{nvf} formulated with barycentric coordinates vs. Euclidean distances. As illustrated in \cref{fig:baryvsspacial}, computing nearest vertices naively by Euclidean distance fails to respect mesh connectivity, particularly for structures like those in \cref{fig:baryvsspacial}. 

\begin{figure}[t]
  \centering
  \sffamily
  \def\svgwidth{\linewidth}
  %% Creator: Inkscape 1.4.2 (f4327f4, 2025-05-13), www.inkscape.org
%% Accompanies image file 'ablation.pdf' (pdf, eps, ps)
\begingroup%
  \makeatletter%
  \providecommand\color[2][]{%
    \errmessage{(Inkscape) Color is used for the text in Inkscape, but the package 'color.sty' is not loaded}%
    \renewcommand\color[2][]{}%
  }%
  \providecommand\transparent[1]{%
    \errmessage{(Inkscape) Transparency is used (non-zero) for the text in Inkscape, but the package 'transparent.sty' is not loaded}%
    \renewcommand\transparent[1]{}%
  }%
  \providecommand\rotatebox[2]{#2}%
  \newcommand*\fsize{\dimexpr\f@size pt\relax}%
  \newcommand*\lineheight[1]{\fontsize{\fsize}{#1\fsize}\selectfont}%
  \ifx\svgwidth\undefined%
    \setlength{\unitlength}{347bp}%
    \ifx\svgscale\undefined%
      \relax%
    \else%
      \setlength{\unitlength}{\unitlength * \real{\svgscale}}%
    \fi%
  \else%
    \setlength{\unitlength}{\svgwidth}%
  \fi%
  \global\let\svgwidth\undefined%
  \global\let\svgscale\undefined%
  \makeatother%
  \begin{picture}(1,0.2353438)%
    \lineheight{1}%
    \setlength\tabcolsep{0pt}%
    \put(0,0){\includegraphics[width=\unitlength,page=1]{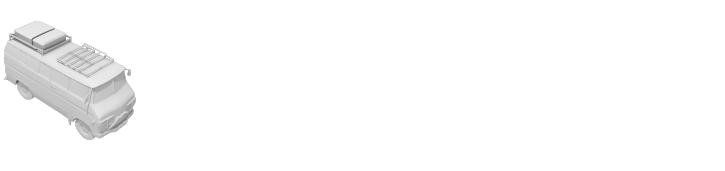}}%
    \put(0.10767217,0.00503373){\makebox(0,0)[t]{\lineheight{1.2643069}\smash{\begin{tabular}[t]{c}\scriptsize Input\end{tabular}}}}%
    \put(0,0){\includegraphics[width=\unitlength,page=2]{ablation.pdf}}%
    \put(0.71356511,0.0048882){\makebox(0,0)[t]{\lineheight{1.2643069}\smash{\begin{tabular}[t]{c}\scriptsize w/o Augment\end{tabular}}}}%
    \put(0,0){\includegraphics[width=\unitlength,page=3]{ablation.pdf}}%
    \put(0.31009635,0.004579){\makebox(0,0)[t]{\lineheight{1.2643069}\smash{\begin{tabular}[t]{c}\scriptsize w/o Watershed\end{tabular}}}}%
    \put(0,0){\includegraphics[width=\unitlength,page=4]{ablation.pdf}}%
    \put(0.51205545,0.00481029){\makebox(0,0)[t]{\lineheight{1.2643069}\smash{\begin{tabular}[t]{c}\scriptsize w/o QEM\end{tabular}}}}%
    \put(0,0){\includegraphics[width=\unitlength,page=5]{ablation.pdf}}%
    \put(0.91551894,0.0039633){\makebox(0,0)[t]{\lineheight{1.2643069}\smash{\begin{tabular}[t]{c}\scriptsize Ours\end{tabular}}}}%
  \end{picture}%
\endgroup%

  \caption{Ablation Study. Results highlight the impact of our three main contributions: watershed algorithm for topological alignment to the prediction; constrained \glsentryshort{qem} for geometry preservation; and data augmentation for robustness.
  }
  \label{fig:ablation}
\end{figure}

\begin{table}[t]
\centering
\caption{Ablations on TRELLIS~\cite{xiang2025trellis} Geometries. The Chamfer Distance is scaled by a factor of $1000$. $\text{M}_{\{\mathcal{G},\mathcal{T}\}}$ represents perceptual scores of geometrical similarity and topological preference to artist meshes provided by \glspl*{vlm}. Removing any key component of our approach results in a performance degradation.}
\label{tab:ablation}
\autoresizetable{
\begin{tabular}{lcccc} 
\toprule
\textbf{Method}  & \textbf{CD}$\downarrow$ & $\textbf{FID}$$\downarrow$ & $\textbf{M}_\mathcal{G}$$\uparrow$ & $\textbf{M}_\mathcal{T}$$\uparrow$  \\ 
\midrule
w/o Watershed    & 0.21                 & 25.95                                & 3.9                                 & \second{3.8}                              \\
w/o QEM          & 0.21                 & 29.13                                & 4.1                                 & 3.5                                       \\
w/o Augmentation & \best{0.20}          & \second{18.06}                       & \second{4.2}                        & 3.7                                       \\
Ours             & \best{0.20}        & \best{16.19}                         & \best{4.4}                          & \best{4.0}                                \\
\bottomrule
\end{tabular}
}
\end{table}

\para{Watershed Grouping Enforces Topology Alignment.}
We ablate the watershed algorithm and apply constrained \gls*{qem} to \gls*{nvf} predictions by directly adding positional quadrics $Q_t$ to all vertex quadrics in \gls*{qem}.
This penalizes large displacements between vertices $\bs{v}_d$ and their predicted target positions $\bs{x}(\bs{v}_d)$ (\cref{sec:extractmesh}). The \gls*{qem} is also performed without rejecting edge collapse across watershed groups.
\cref{tab:ablation} (w/o Watershed) shows performance degradation in geometric and topological quality, and \cref{fig:ablation} visualizes the irregular triangulation,
confirming the utility of the watershed algorithm in enforcing topology alignment to the network prediction.

\para{Constrained QEM Prevents Topological Artifacts and Enforces Geometry Preservation.}
Replacing the constrained \gls*{qem} optimization with a naive iterative vertex-flow (driven solely by the \gls*{nvf}) results in significant topological artifacts and discontinuities (\cref{tab:ablation} w/o QEM). As shown in \cref{fig:ablation}, the absence of geometric constraints leads to ``missing'' geometry, highlighting the necessity of geometric constraints during mesh extraction.

\para{Geometric Data Augmentation Enables Robust Generalization.}
Without the distortion augmentation, the trained model fails to generalize to local surface variations (\cref{tab:ablation,fig:ablation}, w/o Augmentation), leading to a failed topology prediction in those regions.

\para{Limitations.}
The primary limitation of our approach is the reliance on a voxelized representation, which poses challenges for scaling to large-scale scenes. Future work could address this by adopting a multi-resolution, coarse-to-fine generation strategy or a chunk-based pipeline to enable efficient artist-like topology generation of expansive 3D environments.
Besides, \name{} focuses on generating triangle meshes only to avoid the planar ambiguities inherent to n-gon faces. Extending the \gls*{nvf} representation to native n-gon (e.g., quad-dominant) topology is a meaningful next step.

\section{Conclusion}
\label{sec:conslusion}

In this paper, we present \name{}, a generative approach that models mesh topology as a \glsentrylong{nvf} to enable  structured mesh generation from geometric observations. By representing topology as a piecewise continuous field rather than a discrete sequence, our approach avoids the high computational costs and error accumulation issues in autoregressive mesh generation methods. Experiments show that \name{} achieves state-of-the-art performance in producing compact, artist-like topology while maintaining high geometric fidelity at low computational cost, demonstrating improved generalization over prior learning-based baselines. We believe this work provides a new perspective on bridging the gap between neural geometry synthesis and the topological requirements in practical 3D creation workflows.

\section*{Acknowledgements}
This work was funded by AUDI AG.
Angela Dai was supported by the ERC Starting Grant SpatialSem (101076253) and Matthias Nießner by  the ERC Consolidator Grant Gen3D (101171131).
We thank Rui Xu for assistance in running MeshMosaic.

\bibliographystyle{splncs04}
\bibliography{main}

\clearpage
\appendix
\renewcommand\thefigure{A.\arabic{figure}}
\renewcommand\thetable{A.\arabic{table}}
\renewcommand\thealgorithm{A.\arabic{algorithm}}
\renewcommand\theequation{A.\arabic{equation}}
\setcounter{figure}{0}
\setcounter{table}{0}
\setcounter{algorithm}{0}
\setcounter{equation}{0}
\section*{Appendix}
In this material, we provide extended technical details, implementation details, and additional experimental results. In \cref{sec:supp_dataaugmentation}, we describe our data filtering pipeline and the mathematical formulation of the distortion fields used for data augmentation. \cref{sec:supp_vecset} demonstrates \name{}'s generalization to meshes produced by vecset-based 3D generators, and \cref{sec:supp_scaling} discusses scaling \name{} to higher voxelization resolutions. Subsequently, \cref{sec:supp_implementation_details} details the implementation in training, mesh extraction, and perceptual study. \cref{sec:supp_experiment} provides additional experimental statistics, including a runtime breakdown and topological validity metrics compared to existing baselines. Finally, \cref{sec:supp_ablation} presents ablation studies regarding our \gls*{nvf} parameterization, the robustness of the root threshold \(\tau\), quad ratio, and the topology weight \(\lambda_t\).

\section{Data Filtering and Augmentation Details}
\label{sec:supp_dataaugmentation}
\para{Data Filtering.}
Although the \gls*{nvf} is defined on the surface of a geometry, the latent generative model
relies on voxelization, which assumes a well-defined intersection-free surface. In practice, this assumption may not hold. Modern 3D
modeling pipelines typically construct assets from multiple parts that are
assembled together, leading to triangle intersections or overlapping
surfaces near part boundaries. Such artifacts introduce ambiguities during
voxelization and are undesirable for training. To mitigate this issue, we
decompose each object into connected components and treat each component as
an independent training sample. 

Since Objaverse contains many non-artist-created meshes, we apply additional filtering to maximize the  training data quality. 
First, we collapse all mesh edges shorter than the voxel diagonal; samples requiring more than $50\%$ triangle reduction during this process are discarded to ensure topological integrity. 
To exclude excessively thin or elongated geometries, we remove meshes with fewer than $40$k occupied voxels at the resolution of $512^3$. 
Furthermore, we filter for mesh regularity by discarding samples with a quad face ratio below $50\%$. 
Since Objaverse contains diverse mesh types (e.g., polygonal and triangulated), we unify this metric by first triangulating all meshes and then using Meshiki~\cite{meshiki} to estimate the proportion of triangle pairs forming quads. 
In this context, the quad ratio serves as a proxy for the regularity of triangle alignment rather than a precise count of polygonal faces.

\para{Data Augmentation.}
Topology generation is frequently applied to imperfect geometry containing
local distortions, as commonly observed in scanned or generated
meshes. To improve robustness, we apply a random
distortion field \(\bs{\delta} : \mathbb{R}^3 \rightarrow \mathbb{R}^3\) that
defines a continuous deformation in 3D space. For a vertex \(\bs{p}\), the
displacement is given by
\begin{equation}
\bs{\delta}(\bs{p}) =
\left[ 1 - \exp\left(- \frac{\|\bs{p} - \bs{v}_{n}\|_2^2}{2 r_s^2} \right) \right]
\sum_{i=1}^{M} w_i\left(\frac{\|\bs{p}-\bs{a}_i\|_2}{r_a}\right) \, \bs{d}_i,
\end{equation}
where \(\bs{v}_{n}\) is the closest vertex of the original mesh to
\(\bs{p}\), \(r_s\) is a static falloff radius controlling how strongly
regions near the original mesh vertices are preserved, and
\(r_a\) is an anchor radius that determines the spatial extent of its influence. \(\{\bs{a}_i\}_{i=1}^M\) are
control points with associated displacements \(\{\bs{d}_i\}_{i=1}^M\), and
\(w_i(r)\) is the Wendland C2 weighting function:
\begin{equation}
w_i(r) =
\begin{cases}
(1-r)^4 (4r + 1), & r < 1, \\
0, & r \ge 1.
\end{cases}
\end{equation}
The distorted \gls*{nvf} is defined consistently by applying the deformation
to both query points and their corresponding target vertices:
\begin{equation}
\bs{t}_\delta(\delta(\bs{p})) =
\delta(\bs{v}_n) - \delta(\bs{p}).
\end{equation}
In practice, the static falloff scales the displacement according to the proximity
to the vertices of the original mesh. As a result, regions with denser triangles
experience smaller perturbations, whereas sparser regions are distorted more
strongly. This ensures that the model learns to preserve fine geometric details
when generating high-resolution meshes, while being able to simplify geometry
effectively when targeting coarser meshes with fewer triangles.

\begin{algorithm}[tb]
\caption{Priority Watershed \gls*{nvf} Refinement on Dense Mesh
}
\label{alg:watershed}
\begin{algorithmic}[1]
\Require Over-tessellated mesh $\mathcal{M}_d=(\mathcal{V}_d,\mathcal{F}_d)$, predicted \gls*{nvf} $\bs{t}_d(\bs{v}_d)$, threshold $\tau$

\State Initialize all vertices as unlabeled
\State Root initialization $\mathcal{R} = \{\bs{r}\in\mathcal{V}_d \mid \|\bs{t}_d(\bs{r})\|_\infty \le \tau\}$  \Comment{\dcref{eq:roots}{Eq\onedot 6}}
\For{$\bs{r}\in\mathcal{R}$} \Comment{Initialize priority queue}
    \State Push $(0,\bs{r},\bs{r})$
\EndFor
\While{priority queue not empty} \Comment{Iterative region expansion}
    \State Pop $(\text{cost}, \bs{u}, \bs{r})$ with smallest cost
    \If{$\bs{u}$ already labeled}
        \State continue
    \EndIf
    \State Assign vertex $\bs{u}$ to root $\bs{r}$
    \For{$\bs{v} \in \text{Neighbor}(\bs{u}, \mathcal{M}_d)$}
        \If{$\bs{v}$ unlabeled}
            \State Push $(\mathrm{cost}(\bs{v}, \bs{r}),\bs{v},\bs{r})$ \Comment{\dcref{eq:cost}{Eq\onedot 7}}
        \EndIf
    \EndFor
\EndWhile

\State Update \gls*{nvf} \Comment{\dcref{eq:updatenvf}{Eq\onedot 8}}
\end{algorithmic}
\end{algorithm}

\section{Generalization to Vecset Generative Models}
\label{sec:supp_vecset}

\begin{figure}[t]
  \centering
  \sffamily
  \def\svgwidth{0.99\linewidth}
  %% Creator: Inkscape 1.4.2 (f4327f4, 2025-05-13), www.inkscape.org
%% Accompanies image file 'hunyuan.pdf' (pdf, eps, ps)
\begingroup%
  \makeatletter%
  \providecommand\color[2][]{%
    \errmessage{(Inkscape) Color is used for the text in Inkscape, but the package 'color.sty' is not loaded}%
    \renewcommand\color[2][]{}%
  }%
  \providecommand\transparent[1]{%
    \errmessage{(Inkscape) Transparency is used (non-zero) for the text in Inkscape, but the package 'transparent.sty' is not loaded}%
    \renewcommand\transparent[1]{}%
  }%
  \providecommand\rotatebox[2]{#2}%
  \newcommand*\fsize{\dimexpr\f@size pt\relax}%
  \newcommand*\lineheight[1]{\fontsize{\fsize}{#1\fsize}\selectfont}%
  \ifx\svgwidth\undefined%
    \setlength{\unitlength}{355.28771973bp}%
    \ifx\svgscale\undefined%
      \relax%
    \else%
      \setlength{\unitlength}{\unitlength * \real{\svgscale}}%
    \fi%
  \else%
    \setlength{\unitlength}{\svgwidth}%
  \fi%
  \global\let\svgwidth\undefined%
  \global\let\svgscale\undefined%
  \makeatother%
  \begin{picture}(1,0.41597472)%
    \lineheight{1}%
    \setlength\tabcolsep{0pt}%
    \put(0.08783174,0.00547513){\makebox(0,0)[t]{\lineheight{0.27080733}\smash{\begin{tabular}[t]{c}\scriptsize Hunyuan\\\scriptsize (vecset-based)\end{tabular}}}}%
    \put(0.26202203,-0.00704567){\makebox(0,0)[t]{\lineheight{1.26431227}\smash{\begin{tabular}[t]{c}\scriptsize Ours\end{tabular}}}}%
    \put(0,0){\includegraphics[width=\unitlength,page=1]{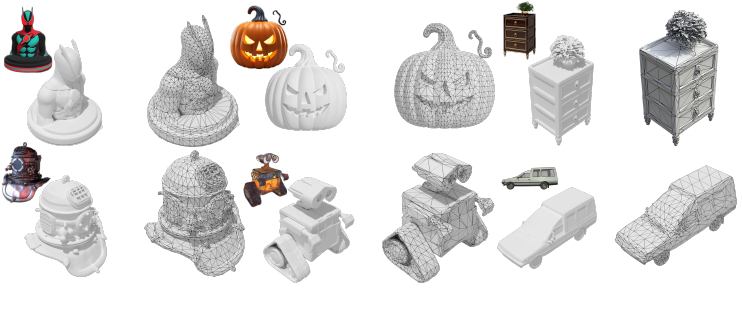}}%
    \put(0.4148736,0.00547513){\makebox(0,0)[t]{\lineheight{0.27080733}\smash{\begin{tabular}[t]{c}\scriptsize Hunyuan\\\scriptsize (vecset-based)\end{tabular}}}}%
    \put(0.58906392,-0.00704567){\makebox(0,0)[t]{\lineheight{1.26431227}\smash{\begin{tabular}[t]{c}\scriptsize Ours\end{tabular}}}}%
    \put(0.74108247,0.00547513){\makebox(0,0)[t]{\lineheight{0.27080733}\smash{\begin{tabular}[t]{c}\scriptsize Hunyuan\\\scriptsize (vecset-based)\end{tabular}}}}%
    \put(0.91610578,-0.00704567){\makebox(0,0)[t]{\lineheight{1.26431227}\smash{\begin{tabular}[t]{c}\scriptsize Ours\end{tabular}}}}%
    \put(0,0){\includegraphics[width=\unitlength,page=2]{hunyuan.pdf}}%
  \end{picture}%
\endgroup%

  \caption{\name{} generalizes to meshes produced by the vecset-based 3D generator Hunyuan3D-2.1~\cite{hunyuan3d2025hunyuan3d}.}
  \label{fig:supp_hunyuan}
\end{figure}

Beyond geometries produced by voxel-based generative models such as TRELLIS~\cite{xiang2025trellis}, we test \name{} on outputs of vecset-based generators. Because \name{} operates on the \gls*{sdf} of its input, applying it to vecset outputs requires no changes to the model. \cref{fig:supp_hunyuan} shows \name{} retopologizing meshes generated by Hunyuan3D-2.1~\cite{hunyuan3d2025hunyuan3d}. \name{} is trained exclusively on Objaverse artist-authored meshes, yet produces compact, artist-like triangle layouts on both voxel- and vecset-generated inputs, suggesting that the \gls*{nvf} representation transfers across upstream generator families.

\section{Scaling to Higher Resolution}
\label{sec:supp_scaling}

\begin{figure}[t]
  \centering
  \sffamily
  \def\svgwidth{\linewidth}
  %% Creator: Inkscape 1.4.2 (f4327f4, 2025-05-13), www.inkscape.org
%% Accompanies image file 'res1024.pdf' (pdf, eps, ps)
\begingroup%
  \makeatletter%
  \providecommand\color[2][]{%
    \errmessage{(Inkscape) Color is used for the text in Inkscape, but the package 'color.sty' is not loaded}%
    \renewcommand\color[2][]{}%
  }%
  \providecommand\transparent[1]{%
    \errmessage{(Inkscape) Transparency is used (non-zero) for the text in Inkscape, but the package 'transparent.sty' is not loaded}%
    \renewcommand\transparent[1]{}%
  }%
  \providecommand\rotatebox[2]{#2}%
  \newcommand*\fsize{\dimexpr\f@size pt\relax}%
  \newcommand*\lineheight[1]{\fontsize{\fsize}{#1\fsize}\selectfont}%
  \ifx\svgwidth\undefined%
    \setlength{\unitlength}{345.0425415bp}%
    \ifx\svgscale\undefined%
      \relax%
    \else%
      \setlength{\unitlength}{\unitlength * \real{\svgscale}}%
    \fi%
  \else%
    \setlength{\unitlength}{\svgwidth}%
  \fi%
  \global\let\svgwidth\undefined%
  \global\let\svgscale\undefined%
  \makeatother%
  \begin{picture}(1,0.38753404)%
    \lineheight{1}%
    \setlength\tabcolsep{0pt}%
    \put(0.12938908,0.00010145){\makebox(0,0)[t]{\lineheight{1.2643044}\smash{\begin{tabular}[t]{c}\scriptsize G.T.\end{tabular}}}}%
    \put(0.38330576,0.00885958){\makebox(0,0)[t]{\lineheight{0.27080733}\smash{\begin{tabular}[t]{c}\scriptsize NVF VAE $512^3$\\\scriptsize Reconstruction\end{tabular}}}}%
    \put(0.63751241,0.00885958){\makebox(0,0)[t]{\lineheight{0.27080733}\smash{\begin{tabular}[t]{c}\scriptsize NVF VAE $1024^3$\\\scriptsize Reconstruction\end{tabular}}}}%
    \put(0.8917191,0.00885958){\makebox(0,0)[t]{\lineheight{0.27080733}\smash{\begin{tabular}[t]{c}\scriptsize Latent Flow $1024^3$\\\scriptsize Generation\end{tabular}}}}%
    \put(0,0){\includegraphics[width=\unitlength,page=1]{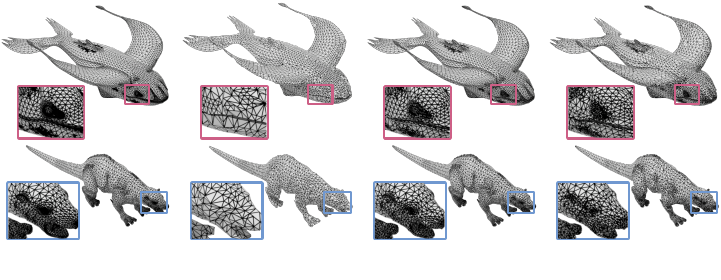}}%
  \end{picture}%
\endgroup%

  \caption{\name{} preserves fine-grained topology at higher resolutions, after fine-tuning on $\sim$150 dense meshes at $1024^3$.}
  \label{fig:supp_res1024}
\end{figure}
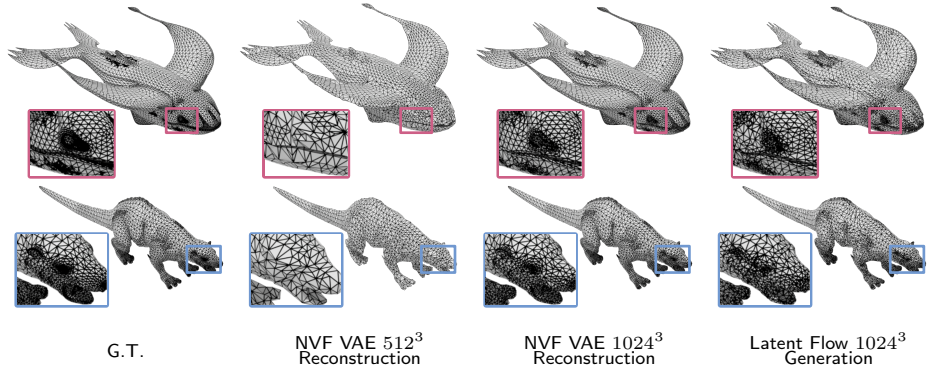

The \gls*{nvf} is defined on the surface and is resolution-independent. In practice, however, our training pipeline voxelizes the field, which sets a lower bound on the size of triangles that can be reliably reconstructed: small features embedded in otherwise smooth regions, such as the whale's eye in \cref{fig:supp_res1024}, collapse when their triangle footprints approach the voxel grid spacing, and the bound recedes proportionally with resolution. Scaling the voxelization from $512^3$ to $1024^3$ halves the spacing and pushes this bound down accordingly, recovering features that were previously unreachable.

The \gls*{nvf} formulation is independent of voxel resolution; the $512^3$ choice in the main paper reflects our training budget rather than a representational limit. As a proof of concept, we fine-tune the \gls*{nvf} \gls*{vae} and the flow model to $1024^3$ using the same positional-encoding-frequency-rescaling technique adopted by TRELLIS~\cite{xiang2025trellis} and Direct3D-S2~\cite{wu2025direct3ds2} for high-resolution training. On a held-out set of $\sim$150 dense Objaverse meshes, the fine-tuned $1024^3$ \gls*{nvf} \gls*{vae} achieves a mean directional cosine similarity of $0.996$ and an $\ell_2$ error of $5.46\times10^{-4}$, compared to $0.888$ and $2.41\times10^{-3}$ for the original $512^3$ \gls*{nvf} \gls*{vae} evaluated on the same set. \cref{fig:supp_res1024} shows representative qualitative results: \name{} preserves fine-grained topology that is unreachable at $512^3$.

\section{Additional Implementation Details}
\label{sec:supp_implementation_details}

\para{Resolving Ambiguity of the NVF Definition.}
The \gls*{nvf} is ambiguous at surface points whose largest barycentric weight is tied between two or more vertices; these points lie exactly on the boundaries between the colored regions in \dcref{fig:nvf_surface}{Fig.~3b}. This ambiguous set has measure zero on the surface and does not affect training, since the field is voxelized before it is learned.

\para{Training.}
The \gls*{nvf} \gls*{vae} is adapted from Direct3D-S2's \gls*{sdf} \gls*{vae}~\cite{wu2025direct3ds2}. The decoder prunes empty regions using the sparse voxel coordinates of the input after each up-sampling layer, allowing memory-efficient training.
We change the hidden dimensions in the decoder from $(512, 128, 64, 32)$ to $(1024, 512, 256, 256)$ to capture the details of the \gls*{nvf}. 

During training, low- and high-frequency displacement fields are applied on meshes normalized to a unit cube. Low-frequency distortion uses 32 anchor points with $r_s=0.5$ with displacements of length $\nicefrac{1}{1024}$. High-frequency distortion uses 64 anchors with $r_s=\nicefrac{1}{8}$ with displacements of length $\nicefrac{1}{512}$. The $r_a$ is set to $\nicefrac{1}{8}$.

\para{Implementation.}
Dataset filtering, pre-processing, and \gls*{nvf} computation are implemented using Trimesh~\cite{trimesh}, MeshLib~\cite{meshlib2025}, Open3D~\cite{zhou2018open3d}, and Meshiki~\cite{meshiki}. The \gls*{nvf} post-processing, including bilateral filtering and the watershed algorithm, are implemented in Python, while the constrained \gls*{qem} is implemented based on Fast-QEM~\cite{pyfqmr,forstmann2014fastqem} in C++. We provide the details of the watershed algorithm in \cref{alg:watershed}.

In practice, we relax the constraint in the \gls*{qem} by allowing contractions between nearby vertices within a small spatial threshold, even when their roots differ, provided that the operation does not split an existing region into multiple disconnected components. Furthermore, collapses that introduce non-manifold edges are rejected to maintain mesh validity and to prevent root vertices from becoming disconnected from the mesh.

\para{Perceptual Study.}
The perceptual study for both \glspl*{vlm} and human participants follows the same setting. In the unary study, participants are shown wireframe renderings of each method and asked to rate the geometric similarity and the topological quality on a scale of $1$ (poor) to $5$ (excellent). In the binary study, participants are shown the results of two methods side-by-side and asked to independently choose the one with better geometry preservation, higher topology regularity, and overall preference.
The \glsentryshort{vlm} evaluations are conducted using Qwen-3.5-Plus~\cite{qwen35blog} and Gemini-3-Flash~\cite{team2023gemini} with around $1700$ queries in total.
Furthermore, we collected $357$ effective human responses for the unary perceptual study and $622$ for the binary perceptual study from $33$ human participants. We observe that the human evaluations align closely with \glspl*{vlm}, as shown in \dcref{tab:comparison,tab:preference}{Tabs\onedot 1 and 2}, highlighting the reliability of both evaluations.

Our \glsentryshort{vlm} prompting adapts the GPTEval3D~\cite{wu2024gpteval3d} to mesh wireframe comparisons, with method order randomized across queries to avoid positional bias. Full prompts are provided in Prompts~\ref{prompt:vlm_unary} and~\ref{prompt:vlm_binary}.

\begin{table}[tb]
\centering
\caption{Additional Statistics. NE stands for the ratio of non-manifold edges. Compared to learning-based methods~\cite{xu2025meshmosaic,lionar2025treemeshgpt}, our method has significantly reduced topological error and fast inference, achieving topology validity and speed comparable to classical baselines~\cite{garland1997qem,huang2018quadriflow}.}
\label{tab:additional_statistics}
\autoresizetable{ 
\begin{tblr}{
  colspec = {l | *{4}{c} | *{4}{c}},
  row{1-2} = {font=\bfseries}, 
  column{1} = {font=\bfseries},
}
\toprule
\SetCell[r=2]{} Method & \SetCell[c=4]{c} Objaverse & & & & \SetCell[c=4]{c} TRELLIS & & & \\
Method & NE$\downarrow$ & \#Face & \#Components$\downarrow$ & Time/s$\downarrow$ & NE$\downarrow$ & \#Face & \#Components$\downarrow$ & Time/s$\downarrow$ \\
\midrule
QEM~\cite{garland1997qem}         & 0.3\textperthousand & 1132    & 1.1 & 22             & 0.1\textperthousand          & 6154  & \best{1.0} & \best{26}\\
QuadriFlow~\cite{huang2018quadriflow}  & \best{0.1}\textperthousand & 833          & \best{1.0} & 20                   & 0.1\textperthousand          & 4610  & 2.5 & 31         \\
MeshMosaic~\cite{xu2025meshmosaic}  & 8.6\textperthousand        & 1563           & 51.6 & 66                  & 8.7\textperthousand          & 95297 & 1493.8 & 7813       \\
TreeMeshGPT~\cite{lionar2025treemeshgpt} & 1.0\textperthousand        & 1606           & 5.6 & 57                    & 4.0\textperthousand          & 6999  & 36.1 & 258        \\
Ours        & \best{0.1\textperthousand}      & \second{1146}  & \second{1.1}& \best{15}  & \best{0.0\textperthousand}   & 6154  & \best{1.0} & 31         \\
\bottomrule
\end{tblr}
}
\end{table}

\begin{table}[t]
\centering
\caption{Ablation on \glsentryshort{nvf} parameterizations. All models are trained for 20k iterations, showing that the performance difference is already distinguishable at the early stage of training. The reported metrics include the synthesized \gls*{nvf}'s $\ell_2$ distance $\ell_2(\bs{t})$, directional cosine similarity $\cos_{\bs{d}}$,  $\ell_1$ error of the length $\ell_1(\|\bs{t}\|_2)$, and $\ell_2$ distance of target positions $\ell_2(\bs{x})$, which shows numerically the accuracy of the prediction. Our parameterization using direction and square root length outperforms other parameterizations.}
\label{tab:additional_ablation}
\autoresizetable{
\begin{tabular}{lcccc} 
\toprule
\textbf{Method}  & $\bs{\ell}_2(\bs{t})$$\downarrow$ & $\bs{\cos}_{\bs{d}}$$\uparrow$ & $\bs{\ell}_1(\|\bs{t}\|_2)$$\downarrow$ & $\bs{\ell}_2(\bs{x})$$\downarrow$  \\ 
\midrule
Parameterize w/ vectors          & 0.14                 & 0.17                                & 0.07                                 & 0.14                                      \\
Parameterize w/ target positions & 0.16          & 0.16                       & 0.09                        & 0.14                                      \\
Ours              & \best{0.10}        & \best{0.26}                         & \best{0.06}                          & \best{0.11}                                \\
\bottomrule
\end{tabular}
}
\end{table}

\section{Additional Experimental Statistics}
\label{sec:supp_experiment}

In \dcref{sec:experiment_comparison}{Sec\onedot 4.4}, we report perceptual ratings and preference on the output meshes, which reflect the visual geometry similarities of the shaded renderings and regularities of the wireframe renderings in human preference. We further report topological statistics, including the ratio of non-manifold edges, the number of faces, and the number of disconnected components, along with runtime, in \cref{tab:additional_statistics}, demonstrating the topological validity and computational efficiency of these methods. The result demonstrates that \name{} outperforms learning-based baselines~\cite{lionar2025treemeshgpt,xu2025meshmosaic} by achieving fewer non-manifold edges and components, as well as more than $8\times$ inference speedup. Generally, \name{} has the same performance level as classical baselines~\cite{garland1997qem,huang2018quadriflow} but generates artist-like topology that is better suited for production (\dcref{sec:experiment_comparison}{Sec\onedot 4.4}). By defining the topology as \gls*{nvf}, \name{} not only achieves higher generation speed, but also benefits from existing optimization paradigms in \gls*{qem} with fewer topological errors.

\para{Control Precision.}
\name{}'s face-count and quad-ratio inputs are conditioning signals that shape the generated field rather than hard constraints. On $100$ Objaverse test samples targeting $1{,}000$ to $8{,}000$ faces, the face count tracks the target with a mean relative error of $-2.01\%$ (variance $3.49\%$), and the quad ratio with a mean error of $-0.21$ (variance $0.01$). The face-count condition drives the level-of-detail behavior (\dcref{fig:lodresult}{Fig.~5}), while the quad-ratio condition biases the field away from degenerate topology (\cref{fig:ablation_quad_ratio}).

\para{Runtime breakdown.}
On average, the flow-matching procedure, involving $50$ denosing steps, occupies $22.8\%$ of the processing time. The bilateral filtering takes $12.7\%$. The watershed algorithm takes $37.3\%$. Finally, the constrained \gls*{qem} takes $18.7\%$. The rest $8.5\%$ is mainly the cost of memory allocations for the \gls*{nvf}. Note that we adapt Fast-QEM~\cite{forstmann2014fastqem} in our implementation, which compromises accuracy for speed. Therefore, the runtime of our \gls*{qem} is lower than the baseline \gls*{qem}, which has higher accuracy. Our bilateral and watershed methods are implemented in Python, which are not fully optimized for speed. Further improvement with parallel processing is an interesting future avenue to explore.

\section{Additional Ablations}
\label{sec:supp_ablation}

\para{Our parameterization of the \gls*{nvf} enables efficient learning.}
We highlight in \dcref{sec:latentflow}{Sec\onedot 3.2} that we parameterize the \gls*{nvf} as $(\bs{d}, s)$, which is the direction and square root of the length of the vector. We conduct an ablation study with two variants: using directly the \gls*{nvf} $\bs{t}(\bs{v})$, and using target positions $\bs{x}(\bs{v}) = \bs{v} + \bs{t}(\bs{v})$. 
We report the synthesized \gls*{nvf}'s $\ell_2$ distance $\ell_2(\bs{t})$, directional cosine similarity $\cos_{\bs{d}}$,  $\ell_1$ error of the length $\ell_1(\|\bs{t}\|_2)$, and $\ell_2$ distance of target positions $\ell_2(\bs{x})$. 
The networks are flow-matching models (w/o latent) trained at a resolution of $128^3$.
As shown in \cref{tab:additional_ablation}, our chosen parameterization achieves significantly improved prediction accuracy, already in the early stage of the training.

\begin{figure}[t]
  \centering
  \sffamily
  \def\svgwidth{0.75\linewidth}
  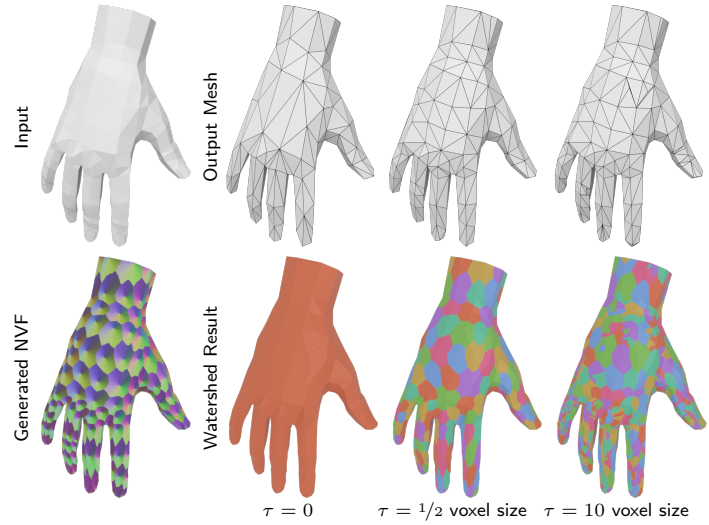
  \caption{Ablation study on root threshold $\tau$. A low threshold clusters the entire surface into a single group, causing the mesh topology to diverge from the generated \glsentryshort{nvf}. Conversely, a high threshold creates redundant surface regions where multiple clusters share the same target vertex. By allowing the \glsentryshort{qem} simplification to merge spatially close vertices from different clusters, the process becomes more robust to variations in $\tau$. Even at a high value of $\tau = 10$ voxel size, the resulting topology remains largely aligned with the \glsentryshort{nvf} prediction.
  }
  \label{fig:ablation_tau}
\end{figure}

\para{Our mesh extraction is robust to variations in the root threshold $\tau$.} 
In \cref{fig:ablation_tau}, we demonstrate the impact of different $\tau$ values during watershed root initialization. 
When $\tau$ is set too low, essential roots are not identified; in the extreme case of $\tau=0$, the entire surface area collapses into a single cluster, failing to reflect the topology implicit in the \gls*{nvf}. 
Conversely, a high $\tau$ value over-segments the surface into redundant subgroups that share the same target vertex. 
However, by allowing the merging of spatially proximal vertices across different clusters (see \cref{sec:supp_implementation_details}), the pipeline remains robust to this over-segmentation. 
As shown in \cref{fig:ablation_tau}, the resulting mesh closely matches the \gls*{nvf} even in the extreme case of $\tau = 10$ voxel size. 
In practice, we recommend $\tau = \nicefrac{1}{2}$ voxel size, or slightly larger, to ensure all primary groups are captured while minimizing unnecessary subdivisions.

\begin{figure}[t]
  \centering
  \sffamily
  \def\svgwidth{0.65\linewidth}
  %% Creator: Inkscape 1.4.2 (f4327f4, 2025-05-13), www.inkscape.org
%% Accompanies image file 'quad_ratio.pdf' (pdf, eps, ps)
\begingroup%
  \makeatletter%
  \providecommand\color[2][]{%
    \errmessage{(Inkscape) Color is used for the text in Inkscape, but the package 'color.sty' is not loaded}%
    \renewcommand\color[2][]{}%
  }%
  \providecommand\transparent[1]{%
    \errmessage{(Inkscape) Transparency is used (non-zero) for the text in Inkscape, but the package 'transparent.sty' is not loaded}%
    \renewcommand\transparent[1]{}%
  }%
  \providecommand\rotatebox[2]{#2}%
  \newcommand*\fsize{\dimexpr\f@size pt\relax}%
  \newcommand*\lineheight[1]{\fontsize{\fsize}{#1\fsize}\selectfont}%
  \ifx\svgwidth\undefined%
    \setlength{\unitlength}{249.31253052bp}%
    \ifx\svgscale\undefined%
      \relax%
    \else%
      \setlength{\unitlength}{\unitlength * \real{\svgscale}}%
    \fi%
  \else%
    \setlength{\unitlength}{\svgwidth}%
  \fi%
  \global\let\svgwidth\undefined%
  \global\let\svgscale\undefined%
  \makeatother%
  \begin{picture}(1,0.29170759)%
    \lineheight{1}%
    \setlength\tabcolsep{0pt}%
    \put(0,0){\includegraphics[width=\unitlength,page=1]{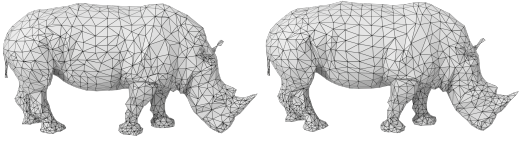}}%
    \put(0.25024701,-0.00873604){\makebox(0,0)[t]{\lineheight{1.2643069}\smash{\begin{tabular}[t]{c}\scriptsize Target Quad Ratio $0.5$\end{tabular}}}}%
    \put(0.75332274,-0.0046016){\makebox(0,0)[t]{\lineheight{1.2643069}\smash{\begin{tabular}[t]{c}\scriptsize Target Quad Ratio $1.0$\end{tabular}}}}%
    \put(0,0){\includegraphics[width=\unitlength,page=2]{quad_ratio.pdf}}%
  \end{picture}%
\endgroup%

  \caption{Ablation study on target quad ratio. Reducing the target quad ratio results in a less regular triangle arrangement, leading to fewer triangle pairs that form quads.
  }
  \label{fig:ablation_quad_ratio}
\end{figure}

\para{The quad ratio acts as an effective filter for irregular topology.} 
As discussed in \cref{sec:supp_dataaugmentation}, the quad ratio serves as a proxy for the regularity of triangle arrangements. 
Consequently, enforcing a higher quad ratio leads to more structured and regular mesh topologies, as illustrated in \cref{fig:ablation_quad_ratio}. 
This condition serves as a sink for poor topology, allowing the model to be trained on a larger variety of samples without being affected by their topological defects. 

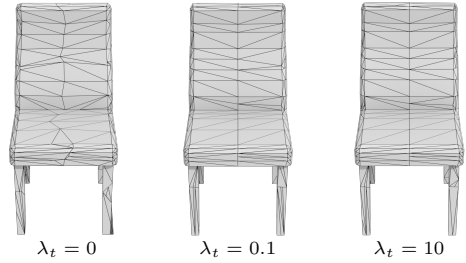
\begin{figure}[t]
  \centering
  \sffamily
  \def\svgwidth{0.5\linewidth}
  %% Creator: Inkscape 1.4.2 (f4327f4, 2025-05-13), www.inkscape.org
%% Accompanies image file 'topology_weight.pdf' (pdf, eps, ps)
\begingroup%
  \makeatletter%
  \providecommand\color[2][]{%
    \errmessage{(Inkscape) Color is used for the text in Inkscape, but the package 'color.sty' is not loaded}%
    \renewcommand\color[2][]{}%
  }%
  \providecommand\transparent[1]{%
    \errmessage{(Inkscape) Transparency is used (non-zero) for the text in Inkscape, but the package 'transparent.sty' is not loaded}%
    \renewcommand\transparent[1]{}%
  }%
  \providecommand\rotatebox[2]{#2}%
  \newcommand*\fsize{\dimexpr\f@size pt\relax}%
  \newcommand*\lineheight[1]{\fontsize{\fsize}{#1\fsize}\selectfont}%
  \ifx\svgwidth\undefined%
    \setlength{\unitlength}{211.0625bp}%
    \ifx\svgscale\undefined%
      \relax%
    \else%
      \setlength{\unitlength}{\unitlength * \real{\svgscale}}%
    \fi%
  \else%
    \setlength{\unitlength}{\svgwidth}%
  \fi%
  \global\let\svgwidth\undefined%
  \global\let\svgscale\undefined%
  \makeatother%
  \begin{picture}(1,0.54711925)%
    \lineheight{1}%
    \setlength\tabcolsep{0pt}%
    \put(0,0){\includegraphics[width=\unitlength,page=1]{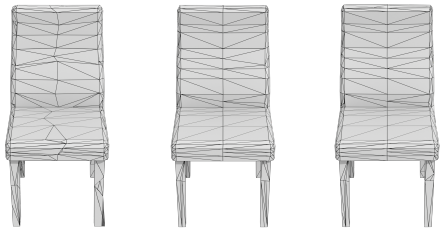}}%
    \put(0.13385681,-0.00702461){\makebox(0,0)[t]{\lineheight{1.2643069}\smash{\begin{tabular}[t]{c}\scriptsize $\lambda_t=0$\end{tabular}}}}%
    \put(0.50515758,-0.00702461){\makebox(0,0)[t]{\lineheight{1.2643069}\smash{\begin{tabular}[t]{c}\scriptsize $\lambda_t=0.1$\end{tabular}}}}%
    \put(0.87645839,-0.00702461){\makebox(0,0)[t]{\lineheight{1.2643069}\smash{\begin{tabular}[t]{c}\scriptsize $\lambda_t=10$\end{tabular}}}}%
  \end{picture}%
\endgroup%

  \caption{Ablation study on topology weight $\lambda_t$. With $\lambda_t=0$, the contraction points are not aligned to the \glsentryshort{nvf} prediction, resulting in undesired vertex placements. With a small $\lambda_t=0.1$, the vertices align with the centerline of the chair, which is identical to an intense weight $\lambda_t=10$. 
  }
  \label{fig:ablation_lambdat}
\end{figure}

\para{A small topology weight $\lambda_t$ effectively snaps vertices to target positions.} 
As illustrated in \cref{fig:ablation_lambdat}, even a small weight of $\lambda_t=0.1$ is sufficient to align the output mesh vertices with the structural features, such as the centerline of the chair. 
This parameter exhibits significant robustness, as the output mesh remains topologically reasonable even when the weight is increased to $\lambda_t=10$. 
In our experiments, we set $\lambda_t=0.1$; this value effectively enforces alignment to the \gls*{nvf} without overpowering the geometric fidelity constraints from the vanilla \gls{qem} formulation.

\refstepcounter{prompt}\label{prompt:vlm_unary}%
\begin{promptbox}{Unary VLM evaluation prompt}
Task Description

You are an expert 3D Technical Artist. Your task is to evaluate 3D
reconstruction methods by comparing their output meshes against a
reference input (far left).

Instructions

Scope: Identify the input and all present reconstruction methods (A, B, C...).

Observation: Closely examine the wireframe. Do not be misled by
low-polygon counts; look for the logic of the edge placement.

Independence of Criteria: Treat Geometry and Topology as separate scores.
A model can have perfect shape but terrible topology, or vice versa.

Review the input mesh: When you consider the method, you need to look
again at the input mesh to have better and consistent scoring. This also
means that the order of the methods shouldn't affect the scoring and you
should not consider other methods when grading the current method.

Evaluation Criteria

CRITERIA 1: SHAPE PRESERVATION AND STRUCTURAL INTEGRITY

Score 1 (Failure):
The mesh exhibits major structural failure. There are significant holes,
often indicated by dark or black rendering artifacts where light cannot
calculate correctly. The geometry is likely non-manifold, missing entire
faces, or consists of disconnected components that should be solid.

Score 2 (Poor):
There is significant loss of volume or silhouette. Critical thin
structures, such as pipes, railings, or antennas, are deleted, severely
broken, or floating as disconnected debris. Large gaps may be present in
the mesh, preventing it from being watertight.

Score 3 (Average):
The general volume of the object is preserved, but thin structures
appear crumbly, jagged, or distorted. Surface continuity is acceptable,
but sharp features are often over-smoothed or eroded. The object looks
like a melted version of the input.

Score 4 (Good):
The mesh is watertight with no dark hole artifacts. Most features are
preserved. There is minor simplification of thin structures, but they
remain connected and coherent. The silhouette matches the input closely.

Score 5 (Excellent):
The mesh is perfectly watertight and structurally complete. Thin
structures like cables, railings, or mechanical parts are distinct,
straight, and continuous. Sharp hard-surface features are crisp.
Simplification occurs only on redundant flat areas, never compromising
the structural silhouette.

CRITERIA 2: TOPOLOGY LIKENESS TO ARTIST/CAD MESHES

Note on silver triangles: silver triangles (long, thin triangles) are
characteristic of CAD-style hard-surface meshing and are a positive
signal on those subjects; on organic subjects, they should not be
expected.

Score 1 (Unusable):
The topology is unusable for professional workflows. It is either a
hyper-dense uniform grid (voxel-style) that is impossible to edit, or it
creates broken, non-manifold geometry with inverted normals.

Score 2 (Chaotic):
The mesh features randomized triangulation. Edges do not follow the
geometry's feature lines, resulting in jagged edges on what should be
straight objects. Random star poles and noisy clusters of triangles
appear on flat surfaces without logic.

Score 3 (Automated Decimation):
This represents standard automated decimation. While it
reduces polygon count and is technically adaptive (more triangles on
curves than flats), the triangulation is noisy. Edges cross flat planes
at random angles, and there is no attempt to align edges with the
model's axes. It is difficult to UV map or select "loops."

Score 4 (Advanced/Structural Topology):
The method shows intentional edge flow. Beyond just being adaptive, the
edges start to align with the primary feature lines of the object. Flat
areas are mostly clean, and curved sections show the beginnings of
organized "strips" or "rings." It avoids the "shattered glass" look of
basic decimation, though it may still lack the refinement of a CAD-style
hard-surface mesh.

Score 5 (Artist/CAD Standard):
The topology is adaptive and mimics a professional human artist. Flat
planar surfaces are represented by large triangles with intentionally
less compact wireframes to form connected edge loops. Edges align with
the geometry's primary feature lines and the mechanical axes of the
object. For CAD-style hard-surface subjects, silver triangles are a
positive signal of efficient curved-surface representation.

Output Format

Detected Methods: [List all identified methods]

Analysis: Method [Label]:

Shape Preservation: [Brief summary of shape accuracy] Score: [1-5]

Topology: [Note specific areas of "Intentional Flow" (e.g. tracks, beams)
vs "Structural Noise." Evaluate if the mesh respects mechanical axes.]
Score: [1-5]

Summary Table:

| Method | Shape Preservation | Topology |

| :--- | :--- | :--- |

| [Label] | [Score] | [Score] |
\end{promptbox}

\refstepcounter{prompt}\label{prompt:vlm_binary}%
\begin{promptbox}{Binary (pairwise) VLM evaluation prompt}
Our task here is to compare two 3D objects, both generated from the
same input reference mesh.

We want to decide which one is better according to the provided
criteria.

\# Instruction

You will see an image containing renderings of the input reference and
these two 3D objects.

The left part of the image contains a single-view rendering of the
reference shape. The middle part of the image contains a single-view
rendering of 3D object 1, with triangle edges (wireframe) colored
black, and the right part contains a single-view rendering for 3D
object 2, with triangle edges (wireframe) colored black.

We would like to compare these two 3D objects from the following
aspects:

1. Artist-involved Topology. This evaluates how alike the mesh
triangles are to a human-created mesh. Please consider symmetry, edge
flow, and adaptive density. 
A topology featuring an uninorm grid structure should be flagged as non-artistic due to a lack of adaptive density.
CAD converted meshes usually
contain long silver triangles, but should be considered an artist
mesh. Silver triangles are characteristic of CAD-style hard-surface
meshing; on organic subjects, they should not be expected. 
Low-poly meshes should be considered artist meshes and could be
identified by looking at regular regions to see how the edges are
aligned. When you consider topology, you should also pay more attention
to large clean areas where you will see the edges more clearly, due to
the resolution of the image, looking at dense regions might mislead
your judgment.

2. Shape Preservation (Geometry). This evaluates how well the
reconstruction preserves the physical volume, proportions, and sharp
features of the Input. Please first describe each of the two models,
and find all missing geometry and broken surface, then evaluate how
well it covers ALL the structures in the original input shape.

3. Professional Preference. This shows your personal preference as a
professional artist. For each of the object, consider how likely you
will use it as an asset in video games, movie production, and to do
manual editing in professional 3D softwares.

Take a really close look at each of the single-view images for these
two 3D objects and the input mesh before providing your answer.

When evaluating these aspects, focus on one of them and the input mesh
at a time.

Try to make independent decision between these criteria.

\# Output format

To provide an answer, please provide a short analysis for each of the
aforementioned evaluation criteria.

The analysis should be very concise and accurate.

For each of the criteria, you need to make a decision using these three
options:

1. object 1 is better;

2. object 2 is better;

3. Cannot decide.

PLEASE CONSIDER OPTION 3 IF THE DIFFERENCE IS LESS THAN 10%

And then, in the last row, summarize your final decision by
"<option for criteria 1> <option for criteria 2> <option for criteria 3>".

An example output looks like follows:

"
Analysis:

1. Artist-involved Topology (Wireframe Logic): Object 1 xxxx; Object 2 xxxx;

Object 1/2 is better or cannot decide

2. Shape Preservation (Geometry): Object 1 xxxx; Object 2 xxxx;

Object 1/2 is better or cannot decide

3. Professional Preference. Object 1 xxxx; Object 2 xxxx;

Object 1/2 is better or cannot decide

Final answer:

x x x (e.g., 1 2 2 / 3 3 3 / 3 2 2)
"
\end{promptbox}

\end{document}